\documentclass[10pt,twocolumn,letterpaper]{article}

\usepackage{iccv}              
\usepackage[pagebackref,breaklinks,colorlinks,allcolors=iccvblue]{hyperref}
\usepackage[accsupp]{axessibility}
\usepackage{multirow}
\usepackage{colortbl}
\usepackage{array}
\usepackage{float}  
\usepackage{amsmath}
\usepackage{hhline}
\usepackage{pifont}
\usepackage{bbding}

\newcommand{\mypara}[1]{\noindent{\bf {#1}.}}

\newcommand{\codename}{{$\mathsf{LightCity}$}\xspace}

\definecolor{iccvblue}{rgb}{0.21,0.49,0.74}
\definecolor{tabfirst}{rgb}{1,0.809,0.809}
\definecolor{tabsecond}{rgb}{0.98,0.82,0.531}
\definecolor{tabthird}{rgb}{0.988,0.957,0.586}

\def\paperID{4332} 
\def\confName{ICCV}
\def\confYear{2025}

\title{\codename: An Urban Dataset for Outdoor Inverse Rendering and Reconstruction under Multi-illumination Conditions}

\author{
Jingjing Wang$^{1}$ $^{\includegraphics[width=0.4cm]{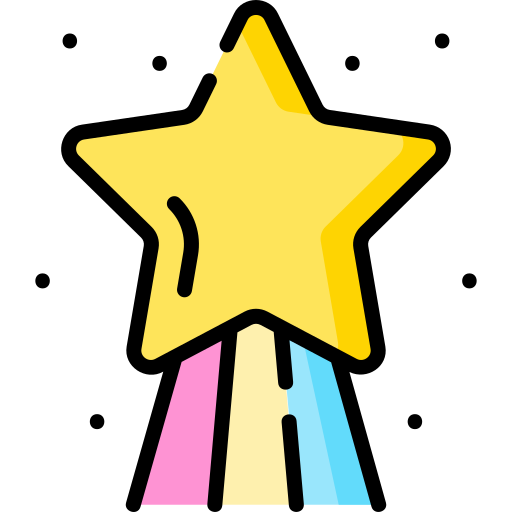}}$\quad
Qirui Hu$^{1}$ $^{\includegraphics[width=0.4cm]{figs/star.png}}$\quad
Chong Bao$^{1}$ $^{\includegraphics[width=0.35cm]{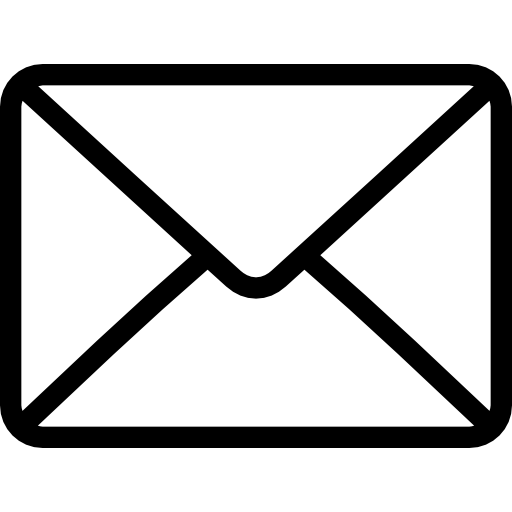}}$  \quad
Yuke Zhu$^{1}$ \\
Hujun Bao$^{1}$ \quad
Zhaopeng Cui$^{1}$ \quad
Guofeng Zhang $^{1}$ $^{\includegraphics[width=0.35cm]{figs/envelope.png}}$ \\
$^{1}$State Key Lab of CAD\&CG, Zhejiang University
}

\newcommand{\ssecspace}{\vspace{-0.5em}}

\newcommand{\secspace}{\vspace{-0.5em}}

\begin{document}

\twocolumn[{%
\renewcommand\twocolumn[1][]{#1}%
\maketitle
\vspace{-3.5em}
\begin{center}
    \centering
    \includegraphics[width=1.0\linewidth, trim={0 0 0 0}, clip]{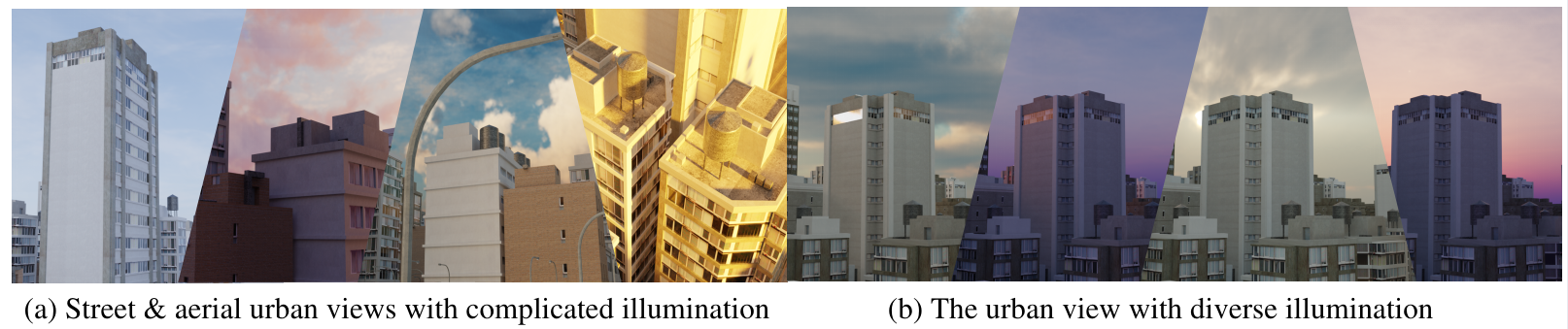}
    \vspace{-2.0em}
    \captionof{figure}{We present a novel high-quality synthetic urban dataset, named \codename. Our dataset features complicated urban illumination conditions, including varied illumination, realistic indirect lighting and shadow effects, realistic indirect light and shadow effects, and varying scales with street and aerial image capture.}
    \label{fig:teaser}
\end{center}%
}]

\renewcommand{\thefootnote}{}
\footnotetext{\includegraphics[width=0.4cm]{figs/star.png} indicates equal contribution.}
\footnotetext{\includegraphics[width=0.35cm]{figs/envelope.png} indicates corresponding author.}
\renewcommand{\thefootnote}{\arabic{footnote}} 

\maketitle
   
\begin{abstract}
Inverse rendering in urban scenes is pivotal for applications like autonomous driving and digital twins. Yet, it faces significant challenges due to complex illumination conditions, including multi-illumination and indirect light and shadow effects.
However, the effects of these challenges on intrinsic decomposition and 3D reconstruction have not been explored due to the lack of appropriate datasets. 
In this paper, we present \codename, a novel high-quality synthetic urban dataset featuring diverse illumination conditions with realistic indirect light and shadow effects.
\codename encompasses over 300 sky maps with highly controllable illumination, varying scales with street-level and aerial perspectives over 50K images, and rich properties such as depth, normal, material components, light and indirect light, etc.
Besides, we leverage \codename to benchmark three fundamental tasks in the urban environments and conduct a comprehensive analysis of these benchmarks, laying a robust foundation for advancing related research. 
Project page: \href{https://23wjj.github.io/LightCity/}{https://23wjj.github.io/LightCity/}.
\end{abstract}   
\vspace{-1.5em}
\section{Introduction}
\label{sec:intro}

Inverse rendering in urban scenes~\cite{zhu2021derendernet, wang2023neural} has become increasingly important for various applications including autonomous driving, digital twin, and urban planning.
However, the complex illumination conditions in urban scenes make this problem particularly challenging.
Two primary challenges stand out:
1) \textbf{Multi-illumination}:
Illumination is an unconstrained component in the urban scene due to its uncontrollability and rapid change over time. 
In everyday scenarios, urban scenes are subject to diverse illumination conditions influenced by factors such as solar position (e.g., sunrise and sunset), weather variations, and seasonal changes.
2) \textbf{Indirect light and shadow}:
The complicated spatial layout between buildings in the urban scene creates pronounced indirect illumination and shadow effects that significantly impact the scene's appearance.

Addressing these challenges requires algorithms that exhibit robust tolerance to such complex illumination effects.
We outline three key functional requirements:
1) The intrinsic decomposition of the urban scene is robust to illumination change.
For example, the decomposed reflectance should be the same across varying illumination conditions of a given scene.
2) The complicated indirect light and shadow effects should be accurately decomposed in the urban scene for inverse rendering.
For example, the indirect light and shadow should be included in the shading component and isolated from the reflectance.
3) Urban scene reconstruction should yield accurate and multi-view consistent results under complicated illumination challenges.

However, these functionalities remain largely unexplored in urban scene analysis due to the lack of appropriate datasets.
Existing datasets with multiple illumination, indirect light and shadow conditions primarily focus on individual objects~\cite{grosse2009ground, jin2023tensoir, kuang2022neroic, kuang2023stanford, liu2023openillumination, toschi2023relight} or small garden environments~\cite{le2021eden}.
The urban datasets~\cite{crandall2012sfm, liao2022kitti, turki2022mega, lin2022capturing, li2023matrixcity} rarely take complicated illumination conditions into account.
For example, MatrixCity~\cite{li2023matrixcity} leverages Unreal Engine to generate urban scenes with illumination controlled solely by adjusting light direction and intensity, resulting in limited diversity that does not reflect real-world conditions.
Datasets like Phototourism~\cite{snavely2006photo} and OMMO~\cite{lu2023large} target 3D reconstruction of individual urban buildings under multiple illumination conditions but their simple spatial layout obliterates complex indirect light and shadow effects.

In this paper, we first propose a novel synthetic urban dataset, named \codename, with diverse illumination conditions and complex indirect light and shadow effects in broader urban contexts.
Second, we benchmark three fundamental tasks under urban scenes to explore three outlined functionalities on \codename.

Specifically, LightCity is distinguished by the following key features as shown in Tab.~\ref{tab:dataset_comparison}:
1) High quality:
LightCity is built upon the SceneCity~\cite{scenecity} add-on with Blender Cycle engine to deliver photo-realistic images with realistic lighting and shadow effects.
2) Rich illumination diversity:
We incorporate over 300 sky maps spanning the entire day, from dawn to night, providing high controllability in illumination through adjustable rotation and intensity of HDRI maps.
3) Scale:  The dataset encompasses synthetic urban blocks of varying scales, with over 30K views for intrinsic tasks and over 20K views for reconstruction tasks images covering both street-level and aerial perspectives.
4) Comprehensive Properties:
The dataset includes multiple attributes that can support various vision tasks, such as depth and normal maps for geometry estimation, diffuse and glossy components for material estimation, etc.

To demonstrate the utility of our dataset on outlined functionalities, we benchmark three tasks.
1) We evaluate image intrinsic decomposition in the urban scene to investigate intrinsic consistency under different illumination conditions.
Our findings indicate that current image intrinsic decomposition models lack intrinsic coherency of a scene under varying illumination conditions.
Models fine-tuned with \codename tend to learn more consistent intrinsic properties regardless of illumination variations and predict a better shading property.
2) We evaluate multi-view inverse rendering in urban scenes with a single unknown illumination to investigate its intrinsic accuracy and multi-view consistency.
Our findings indicate that current multi-view inverse rendering algorithms fall short in urban scene material estimation and are hard to disentangle view-dependent indirect light, and shadow effects.
3) We evaluate neural rendering in urban scenes with various illuminations to investigate the multi-illumination effect on novel view synthesis.
Our findings indicate that approaches employing 3D Gaussian Splatting (3DGS)~\cite{kerbl20233d} deliver superior rendering quality and geometric consistency compared to neural radiance field (NeRF)~\cite{mildenhall2021nerf}. Nevertheless, 3DGS-based methods~\cite{zhang2024GS-W, xu2024wild, kulhanek2024wildgaussians, dahmani2024swag, tang2024nexussplats} manifest appearance variations induced by multiple illuminations as floating artifacts.

Our contributions are summarized as follows:
\begin{itemize}
\item LightCity features over 300 sky maps with highly controllable illumination. It includes urban blocks of varying scales with both street-level and aerial views. Additionally, it provides rich properties like depth, normal, diffuse, and glossy materials to support diverse vision tasks.
\item We benchmark three fundamental tasks in the urban scene on \codename involving intrinsic image decomposition, multi-view inverse rendering, and urban scene reconstruction under multiple illuminations.
\item We perform an in-depth analysis of the benchmarking results and our findings highlight the impact of multiple illumination on intrinsic decomposition consistency, the effect of indirect light and shadow conditions on inverse rendering, and the accuracy of urban scene reconstruction under diverse illumination.
\end{itemize}

\setlength{\tabcolsep}{0.5pt}
\setlength{\abovecaptionskip}{0.cm}
\setlength{\belowcaptionskip}{-0.cm}
\begin{table}
\centering
\small
\resizebox{1.0\linewidth}{!}{
\tabcolsep 3pt

\vspace{-1.0em}
\begin{tabular}{cccccccc}
\toprule
Datasets          & Task    & \#Images & Level        & Src.    & Intri.    & Mat.  & Light                   \\ 
\toprule
OMMO~\cite{lu2023large}              & Rec.    & 14K      & Wild   & R     & \ding{55} & \ding{55} & R                     \\
MatrixCity~\cite{li2023matrixcity}        & Rec.     & 519K     & City         & S & A                          & \ding{51} & ID                         \\
PhotoTourism~\cite{snavely2006photo}      & Rec.     & 30K      & Landmark & R      & \ding{55} & \ding{55} & R                       \\ 
\midrule
NeRF-OSR~\cite{rudnev2022nerf}          & Rel.     & 3K       & Building  & R    & \ding{55} & \ding{55} & R                        \\
OpenIllum~\cite{liu2023openillumination} & Rel.     & 108K     & Object   &       & \ding{55} & \ding{55} & R                       \\ 
\midrule
MIT Intrinsic~\cite{grosse2009ground}    & Dec     & 110      & Object   & R     & ASR                      & \ding{55} & R                       \\
MPI Sintel~\cite{Butler2012mpisintel}        & Dec     & 2.6K     & Movie  & S & A                          & \ding{55} & \ding{55}  \\
CGIntrinsic~\cite{li2018cgintrinsics}      & Dec     & 20K      & Indoor & S & AS                        & \ding{55} & \ding{55}  \\
IIW~\cite{bell2014intrinsic}               & Dec     & 5K       & Indoor & R     & AS                        & \ding{55} & \ding{55}  \\ 
\midrule
Hypersim~\cite{roberts2021hypersim}          & Und     & 82K      & Indoor & S & ASR                      & \ding{55} & \ding{55}  \\
EDEN~\cite{le2021eden}              & Und     & 439K     & Garden & S & AS                        & \ding{55}& HID                  \\ 
\hline\hline
\textbf{Ours}     & Rec+Dec & 50K  & City  & S & ASR                      & \ding{51} & HID                  \\
\bottomrule 
\end{tabular}
}
\vspace{-1.0em}
\captionsetup{skip=10pt}
\caption{Comparison of properties between our \codename dataset with previous datasets. Task: Rec=Reconstruction, Rel=Relighting, Dec=Decomposition, Und=Understanding. Src. (Source): R=Real, S=Synthetic. Intri.: A=Albedo, S=Shading, R=Residual. Light: R=Real, I=Intensity, D=Direction, H=HDRs.}
\label{tab:dataset_comparison}
\vspace{-2.0em}
\end{table}

\vspace{-0.5em}
\section{Related Works}

Recent advances in 3D scene representation, from Neural Radience Field (NeRF)~\cite{mildenhall2021nerf}to 3D Gaussian Splatting (3DGS)~\cite{kerbl20233d,zhai2025splatloc}, have enabled high-quality reconstructions, including at the urban scale~\cite{turki2022mega, liu2024citygaussian}. However, these methods mostly assume idealized lighting conditions, which rarely hold in real-world data collection. NeRF-W~\cite{martin2021nerf} was the first to model outdoor illumination variations, followed by works of NeRF-based Ha-NeRF~
\cite{chen2022hallucinated}, NeuralRecon~\cite{sun2022neural} and CR-NeRF~\cite{yang2023cross}, 3DGS-based Wild-gs~\cite{xu2024wild}, Wild-gaussian~\cite{kulhanek2024wildgaussians}, Gaussian-wild~\cite{zhang2024GS-W} and SWAG~\cite{dahmani2024swag}, all being evaluated under the PhotoTourism dataset. However, the PhotoTourism lacks diverse lighting interactions between densely placed objects and comprehensive ground truth for geometry evaluation, underscoring the need for a more complex and unified benchmark for outdoor reconstruction under multi-illumination constraints.
 
Beyond scene reconstruction, understanding both image and scene intrinsics, e.g., reflectance $A$ and shading $S$, remains an open problem. Intrinsic image decomposition evolved from Lambertian assumptions ($I=A\times S$)~\cite{tappen2005recovering, shen2008intrinsic, bell2014intrinsic, careaga2023intrinsic, meka2021real, das2022pie, li2018cgintrinsics}, to non-Lambertian models ($I=A\times S+R$)~\cite{shi2017learning, yeh2022learning, careagaColorful}, decomposing specular effects as residual $R$. While widely studied, intrinsic image decomposition has been limited to indoor datasets~\cite{bell2014intrinsic, kocsis2024intrinsic}, with outdoor benchmarks~\cite{Butler2012mpisintel, le2021eden}, being either low-quality or overly simplistic. Given the complexity of outdoor multi-illumination effects, a large-scale dataset is needed to support both training and benchmarking. For scene inverse rendering, which shares a similar goal with intrinsic decomposition, aims to estimate scene albedo from multi-view images while often incorporating physically-based rendering (PBR) models to recover material properties (e.g., roughness and metallicity) for future relighting. Existing NeRF-~\cite{rudnev2022nerf,toschi2023relight,ye2023intrinsicnerf} and 3DGS-~\cite{liang2024gs, gao2024relightable} based inverse rendering methods primarily focus on object datasets~\cite{murmann2019dataset, liu2023openillumination}, while urban outdoor inverse rendering~\cite{wang2023neural, lin2025urbanir, zeng2024rgb, pun2023neural} remains largely unexplored due to dataset limitations. More details are in Suppl.~Sec.~1.
\vspace{-0.3em}
\section{LightCity Dataset}
\begin{figure}[!t]
    \centering
    \includegraphics[width=0.9\linewidth]{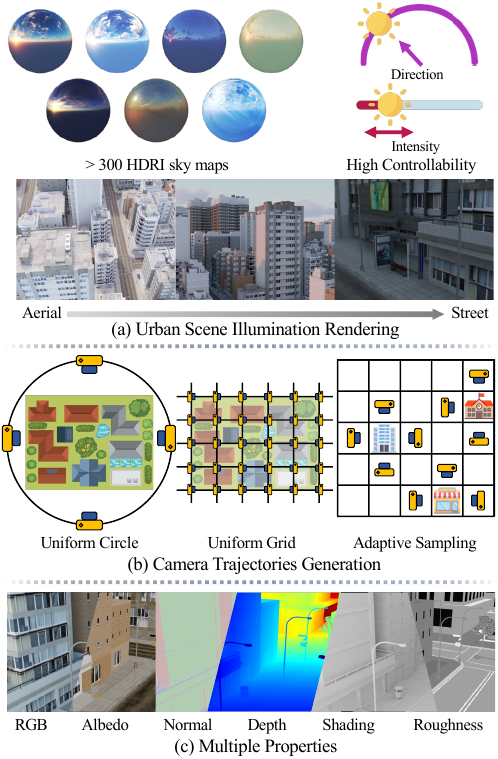}
    \caption{\textbf{Overview.} The features of our urban datasets: (a) Diverse variety and flexible control of illuminations. (b) View sampling with varying scales. (c) Multiple properties.}
    \label{fig:framework}
    \vspace{-0.5cm}
\end{figure}

The \codename dataset aims to provide a challenging benchmark for intrinsic image decomposition, multi-view inverse rendering, and urban scene reconstruction under more realistic multi-illumination constraints.
As illustrated in Fig.~\ref{fig:framework}, the dataset creation follows a three-stage pipeline: collecting and annotating city-scale assets and environment maps (Sec.~\ref{sec:data_collection}), sampling camera views (Sec.~\ref{sec:cam_trajectory}), and rendering various attributes using a PBR approach (Sec.~\ref{sec:rendering}). Built on open-source Blender, our framework is adaptable to various publicly available city assets.  

\subsection{Data Collection and Modeling}
\label{sec:data_collection}
\mypara{City-scale Assets} 
To reveal the multi-illumination challenge in outdoor scenes, our dataset meets three key criteria: high-quality city-scale assets for photorealistic fidelity, diverse object categories with rich color details for comprehensive scene understanding, and a large-scale urban environment with complex object placements to capture long-term illumination dependencies. 
To achieve this, we utilize the SceneCity Blender~\cite{scenecity} add-on, generating a city map with over 450 object categories and 80 material properties, enabling rapid scene asset construction.

\mypara{Block Division and Annotation} 
We organize our base city map into five hierarchical levels, $\{A, B, C, D, E, F\}$ from large to small based on building clusters. Please see Suppl.~Sec.~3 for detailed information.
To annotate the objects within the city map, we begin by decomposing the predefined city assets into individual objects, as each object may be part of a larger asset group. 
Each object is then assigned a semantic label.
 
\mypara{Multi-illumination Modeling} 
To model the diverse illumination conditions for outdoor scenes, we chose the HDRI sky maps as the primary representation. 
Unlike previous synthetic datasets that only adjust light intensity and direction, our approach leverages variations in sky textures, which not only influence illumination but also indicate the time of day. 
We collected over 300 HDRI maps spanning from dawn to night to control global illumination by rotating the HDRI and adjusting the ambient intensity. Compared to other widely used datasets, our dataset exhibits richer dynamic sky variations, introducing more complex lighting challenges for tasks on urban scenes.

\subsection{Camera Trajectories Generation}
\label{sec:cam_trajectory}
\begin{figure}[t]
\centering
\includegraphics[width=0.8\linewidth]{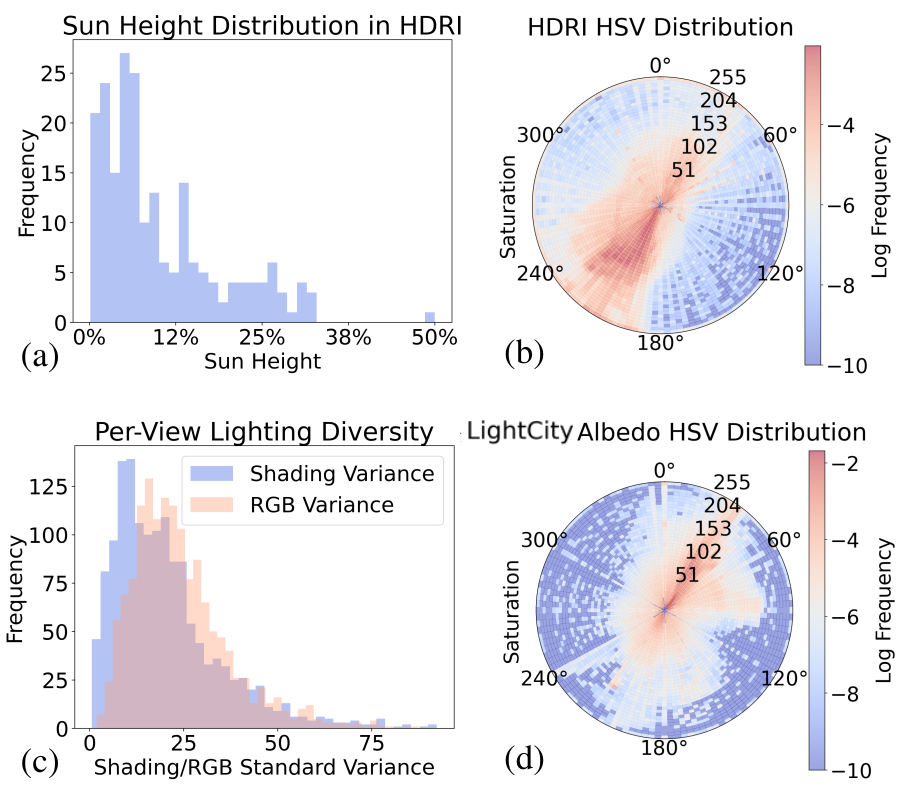}
\caption{(a) The HDRI sun height distribution, with 0\% representing sea level, covers a wide range of the sun's trajectory throughout the day.
(b, d) The HSV distribution of the HDRI maps and albedos spans a wide color space, reflecting diverse lighting and textures.
(c) The distribution of rendered shading and RGB variance shows a wide range of brightness and color changes.}
\label{fig:dataset_distribution}
\vspace{-1.5em}
\end{figure}

\mypara{Uniform View Sampling}
View sampling methods such as the circular sampling and the uniform grid sampling are widely used in previous datasets~\cite{li2023matrixcity, lu2023large,liu2023openillumination}. 
To further eliminate the influence of camera trajectories on outdoor reconstruction quality and better assess the impact of lighting factors on it, we adopted two uniform view sampling methods: uniform circular and uniform grid sampling. 

\mypara{Adaptive View Sampling} 
To construct more detailed and comprehensive viewpoints, we integrate street and aerial views through an adaptive view sampling strategy. Please see Suppl.~Sec.~2 for detailed information.

\mypara{Callibration}
Extensive research has demonstrated that the overlapping regions between camera viewpoints are crucial for high-quality 3D reconstruction. Insufficient overlap between camera views significantly increases the likelihood of reconstruction failure. To refine the randomly generated camera views, we use COLMAP to perform a coarse multi-view point cloud reconstruction and filter out camera poses that fail to establish sufficient feature correspondences.

\subsection{Image Rendering and Filtering}
\label{sec:rendering}
\mypara{Rendering} Building upon the city assets and pre-generated camera viewpoints, we select Blender Cycles, a PBR renderer, as our rendering engine, which efficiently simulates global illumination to achieve photorealistic results. 
PBR relies on BSDF (Bidirectional Scattering Distribution Function) shaders to define material properties such as diffuse reflection, glossy reflection, transparency, etc. It utilizes the G-Buffer to separately store intermediate rendering results, including ambient lighting, geometry, direct and indirect illumination, etc. Following the Blender Cycles Manual~\cite{blender_passes}, a final image $I(\mathbf{x})$ can be defined formally as follows:

\vspace{-1.5em}
\begin{equation}
  \begin{aligned}
I_D(\mathbf{x}) &= g_{D}(\mathbf{x})(g_{Ddirect}(\mathbf{x})+g_{Dindirect}(\mathbf{x})), \\
I_{G}(\mathbf{x}) &= g_{G}(\mathbf{x})(g_{Gdirect}(\mathbf{x})+g_{Gindirect}(\mathbf{x})), \\
I(\mathbf{x}) = &I_{D}(\mathbf{x}) + I_{G}(\mathbf{x}) + I_T(\mathbf{x}) + I_B(\mathbf{x}) + I_E(\mathbf{x}),
  \end{aligned}
  \vspace{-0.5em}
\end{equation}
where $\mathbf{x}$ represents the pixel location, $I_D, I_{G}, I_T, I_B, I_E$ represent the rendered contributions from diffuse BSDF, glossy BSDF, transmission BSDF, background and emission sources. $g_{D(G)}$, $g_{D(G)direct}$ and $g_{D(G)indirect}$ represents the color pass, the direct and indirect lighting pass for the diffuse or glossy BSDF component. To extract image intrinsics from the rendering equation, we formulate the albedo $A(\mathbf{x})$ and shading $S(\mathbf{x})$ as follows:
\vspace{-0.9em}
\begin{equation}
    \begin{aligned}
        A(\mathbf{x}) &= g_{D}(\mathbf{x}), \\
        S(\mathbf{x}) &= g_{Ddirect}(\mathbf{x})+g_{Dindirect}(\mathbf{x}), \\
        R(\mathbf{x}) &= I_{G}(\mathbf{x}) + I_T(\mathbf{x}) + I_B(\mathbf{x}) + I_E(\mathbf{x}).
    \end{aligned}
    \vspace{-0.8em}
\end{equation}
We set up an automatic multi-channel rendering pipeline to generate diverse attributes for our dataset, including albedo, shading, material attributes (i.e., roughness, metallic and specular), normals, semantics, and depth. To better model illumination effects, we randomly select two HDRI maps per camera view and apply four random rotations and ambient intensities.
For a balance between rendering efficiency and visual fidelity, we set the output resolution to 1024×768 and the max render samples to 512, prioritizing photorealism over rendering speed.

\mypara{Filtering} In the final stage of image filtering, we eliminate excessively dark or underexposed images that may negatively impact reconstruction. This is achieved by setting an intensity threshold to remove images with insufficient brightness. Besides, to exclude images with minimal informative content, such as those containing only a single wall, an empty sky, or views extending beyond the city boundaries, we analyze the corresponding semantic maps and filter out images where the total number of distinct objects is fewer than two. Generally, for each image $I$ our filtering criteria can be formulated as follows:
\vspace{-0.5em}
\begin{equation}
    \begin{cases}
    Y_I = 0.299 R_I+ 0.587 G_I+0.114 B_I, Y_I<\tau_{Y},\\
    O_I = \left| \{ l \mid l \in S_I \} \right|, O_I<\tau_{O},
    \end{cases}
    \vspace{-0.5em}
\end{equation}
where $Y_I$, $R/G/B_I$, $S_I$ represents the brightness, single color channel, and semantic map, $\tau_{Y}$ and $\tau_{O}$ represents the threshold set for filtering brightness and semantics.

\setlength{\tabcolsep}{3pt}

\begin{table*}
\centering
\small
\resizebox{1.0\linewidth}{!}{
\tabcolsep 3pt
\arrayrulecolor{black}
\begin{tabular}{c|c|c|ccccc|ccccc} 
\hline
\multicolumn{13}{c}{LightCity-Outdoor}                               \\ 
\hline
\multicolumn{2}{c|}{\multirow{2}{*}{Method}}      & \multirow{2}{*}{$D_{train}$} & \multicolumn{5}{c|}{Albedo}                                                                                                                                                                                & \multicolumn{5}{c}{Shading}                                                                                                                                                                                 \\ 
\cline{4-13}
\multicolumn{2}{c|}{}                             &                     & si-PSNR$\uparrow$                                & SSIM$\uparrow$                                   & LPIPS$\downarrow$                                  & si-MSE$\downarrow$                                 & si-LMSE$\downarrow$                                & si-PSNR$\uparrow$                                   & SSIM$\uparrow$                                   & LPIPS$\downarrow$                                  & si-MSE$\downarrow$                                 & si-LMSE$\downarrow$                                 \\ 
\hline
\multirow{4}{*}{DNN}       & PIE-Net~\cite{das2022pie}              & Outdoor                & 15.49~                                 & {\cellcolor{tabfirst}}0.529~ & 0.449~                                 & 0.053~                                 & 0.050~                                 & 16.15~                                 & 0.463~                                 & 0.570~                                 & 0.114~                                 & 0.106~                                  \\ 
\cline{2-3}
                           & \multirow{2}{*}{DPF~\cite{chen2023dpf}} & H                   & 16.20~                                 & 0.369~                                 & 0.642~                                 & 0.020~                                 & 0.018~                                 & 17.92~                                 & 0.445~                                 & 0.660~                                 & 0.031~                                 & 0.028~                                  \\
                           &                      & H+L                 & {\cellcolor{tabsecond}}18.94~ & {\cellcolor{tabthird}}0.435~ & 0.611~                                 & {\cellcolor{tabsecond}}0.013~ & {\cellcolor{tabsecond}}0.013~ & {\cellcolor{tabsecond}}21.09~ & {\cellcolor{tabthird}}0.496~ & 0.645~                                 & 0.024~                                 & 0.022~                                  \\ 
\hhline{~--~~>{\arrayrulecolor{tabsecond}}-~>{\arrayrulecolor{tabthird}}--~->{\arrayrulecolor{tabsecond}}--}
                           & CDID~\cite{careagaColorful}            & E                 & 15.80~                                 & 0.370~                                 & {\cellcolor{tabsecond}}0.385~ & 0.016~                                 & {\cellcolor{tabthird}}0.015~ & {\cellcolor{tabthird}}20.25~ & 0.204~                                 & {\cellcolor{tabthird}}0.542~ & {\cellcolor{tabsecond}}0.013~ & {\cellcolor{tabsecond}}0.012~  \\ 
\hhline{>{\arrayrulecolor{black}}--->{\arrayrulecolor{tabthird}}-~-->{\arrayrulecolor{tabsecond}}-~-->{\arrayrulecolor{tabthird}}--}
\multirow{3}{*}{Diffusion} & \multirow{2}{*}{DMP~\cite{lee2024exploiting}} & H                   & {\cellcolor{tabthird}}17.61~ & 0.397~                                 & {\cellcolor{tabthird}}0.448~ & {\cellcolor{tabthird}}0.015~ & {\cellcolor{tabsecond}}0.013~ & 18.38~                                 & {\cellcolor{tabsecond}}0.525~ & {\cellcolor{tabsecond}}0.526~ & {\cellcolor{tabthird}}0.023~ & {\cellcolor{tabthird}}0.020~  \\
                           &                      & H+L                 & {\cellcolor{tabfirst}}22.80~ & {\cellcolor{tabsecond}}0.520~ & {\cellcolor{tabfirst}}0.363~ & {\cellcolor{tabfirst}}0.005~ & {\cellcolor{tabfirst}}0.005~ & {\cellcolor{tabfirst}}26.58~ & {\cellcolor{tabfirst}}0.611~ & {\cellcolor{tabfirst}}0.432~ & {\cellcolor{tabfirst}}0.003~ & {\cellcolor{tabfirst}}0.003~  \\ 
\arrayrulecolor{black}\cline{2-3}
                           & IntrinsicAny~\cite{chen2024intrinsicanything}             & Objects                & 16.82~                                 & 0.390~                                 & 0.535~                                 & 0.049~                                 & 0.047~                                 & \multicolumn{5}{c}{/}                                                                                                                                                                                       \\
\hline
\end{tabular}
}
\caption{ Performance comparison of intrinsic decomposition on \codename.  The \colorbox{tabfirst}{first}, \colorbox{tabsecond}{second}, and \colorbox{tabthird}{third} values are highlighted. }
\label{tab: lightcity_intrinsic}
\vspace{-0.8em}
\end{table*}

\setlength{\tabcolsep}{3pt}
\begin{table*}
\centering
\small
\resizebox{1.0\linewidth}{!}{
\tabcolsep 3pt

\arrayrulecolor{black}
\begin{tabular}{c|c|c|ccccc|ccccc} 
\hline
\multicolumn{13}{c}{EDEN-Outdoor OOD}                               \\ 
\hline
\multicolumn{2}{c|}{\multirow{2}{*}{Method}}      & \multirow{2}{*}{$D_{train}$} & \multicolumn{5}{c|}{Albedo}                                                                                                                                                                                & \multicolumn{5}{c}{Shading}                                                                                                                                                                                 \\ 
\cline{4-13}
\multicolumn{2}{c|}{}                             &                     & si-PSNR$\uparrow$                                & SSIM$\uparrow$                                   & LPIPS$\downarrow$                                  & si-MSE$\downarrow$                                 & si-LMSE$\downarrow$                                & si-PSNR$\uparrow$                                   & SSIM$\uparrow$                                   & LPIPS$\downarrow$                                  & si-MSE$\downarrow$                                 & si-LMSE$\downarrow$                                 \\ 
\hline
\multirow{4}{*}{DNN}       & PIE-Net~\cite{das2022pie}              & Outdoor                 & 6.397~                                 & 0.094~                                 & 0.538~                                 & 0.973~                                 & 0.956~                                 & 6.58~                                  & 0.050~                                 & 0.499~                                 & 3.396~                                 & 3.256~                                  \\ 
\hhline{~--~~~~~>{\arrayrulecolor{tabthird}}-~~--}
                           & \multirow{2}{*}{DPF~\cite{chen2023dpf}} & H                   & 11.39~                                 & 0.125~                                 & 0.592~                                 & 0.196~                                 & 0.191~                                 & {\cellcolor{tabthird}}10.56~ & 0.053~                                 & 0.568~                                 & {\cellcolor{tabthird}}1.602~ & {\cellcolor{tabthird}}1.473~  \\
                           &                      & H+L                 & 13.11~                                 & 0.094~                                 & 0.580~                                 & {\cellcolor{tabsecond}}0.159~ & {\cellcolor{tabthird}}0.151~ & {\cellcolor{tabfirst}}14.16~ & {\cellcolor{tabfirst}}0.388~ & {\cellcolor{tabsecond}}0.452~ & {\cellcolor{tabfirst}}0.552~ & {\cellcolor{tabfirst}}0.476~  \\ 
\hhline{~>{\arrayrulecolor{black}}-->{\arrayrulecolor{tabfirst}}-->{\arrayrulecolor{tabsecond}}->{\arrayrulecolor{tabfirst}}--~>{\arrayrulecolor{tabsecond}}-~--}
                           & CDID~\cite{careagaColorful}            & E+Indoor.etc                 & {\cellcolor{tabfirst}}17.54~ & {\cellcolor{tabfirst}}0.234~ & {\cellcolor{tabsecond}}0.351~ & {\cellcolor{tabfirst}}0.057~ & {\cellcolor{tabfirst}}0.055~ & 9.39~                                  & {\cellcolor{tabsecond}}0.131~ & 0.550~                                 & {\cellcolor{tabsecond}}0.662~ & {\cellcolor{tabsecond}}0.646~  \\ 
\hhline{>{\arrayrulecolor{black}}--->{\arrayrulecolor{tabthird}}---~~~~-~~}
\multirow{3}{*}{Diffusion} & \multirow{2}{*}{DMP~\cite{lee2024exploiting}} & H                   & {\cellcolor{tabthird}}13.12~ & {\cellcolor{tabthird}}0.139~ & {\cellcolor{tabthird}}0.382~ & 0.307~                                 & 0.297~                                 & 8.09~                                  & 0.034~                                 & {\cellcolor{tabthird}}0.470~ & 3.760~                                 & 3.482~                                  \\
                           &                      & H+L                 & {\cellcolor{tabsecond}}15.10~ & {\cellcolor{tabsecond}}0.158~ & {\cellcolor{tabfirst}}0.347~ & {\cellcolor{tabthird}}0.163~ & {\cellcolor{tabsecond}}0.150~ & {\cellcolor{tabsecond}}11.54~ & {\cellcolor{tabthird}}0.075~ & {\cellcolor{tabfirst}}0.402~ & 1.766~                                 & 1.572~                                  \\ 
\arrayrulecolor{black}\cline{2-3}
                           & IntrinsicAny~\cite{chen2024intrinsicanything}             & Objects                 & 6.293~                                 & 0.059~                                 & 0.573~                                 & 0.866~                                 & 0.855~                                 & \multicolumn{5}{c}{/}   \\
\hline
\end{tabular}
}
\caption{Performance comparison of intrinsic decomposition on EDEN. The \colorbox{tabfirst}{first},\colorbox{tabsecond}{second}, and \colorbox{tabthird}{third} values are highlighted.}
\label{tab: eden_intrinsic}
\vspace{-2.0em}
\end{table*}

\section{Characteristic Analysis}
In this section, we analyze the statistical characteristics of our proposed \codename dataset. Detailed information on our dataset is provided in Suppl.~Sec.~3.

\mypara{Controllable Multi-illuminations} Our synthetic dataset offers precise control over diverse outdoor lighting conditions, surpassing real-world datasets. Beyond adjusting the direction and intensity of global illumination, \codename focuses on view-dependent sky textures and accurately models ambient urban lighting. Our HDRI maps span from early morning to midnight, capturing a rich color and brightness distribution, as shown in Fig.~\ref{fig:dataset_distribution} (a)(b). This comprehensive control enhances its suitability for urban reconstruction.

\mypara{Diverse Intrinsics} \codename provides an automated multi-attribute rendering pipeline, generating ground-truth labels for PBR-based inverse rendering and intrinsic image decomposition. It provides decomposed lighting components and fundamental material properties such as metallic, roughness, and specular. As shown in Fig.~\ref{fig:dataset_distribution} (c), the distribution of the variance for our rendered RGB images and shading under the same view spans a broad range, indicating a rich light variation. Compared with MatrixCity (provided in Suppl.~Sec.~3), the HSV distribution of our diffuse color favors a broader range of color space (Fig.~\ref{fig:dataset_distribution} (d)). The rich and diverse intrinsics prepare our dataset for a broader range of future applications.

\begin{figure*}[th]
\centering
\tiny
\includegraphics[width=0.8\linewidth]{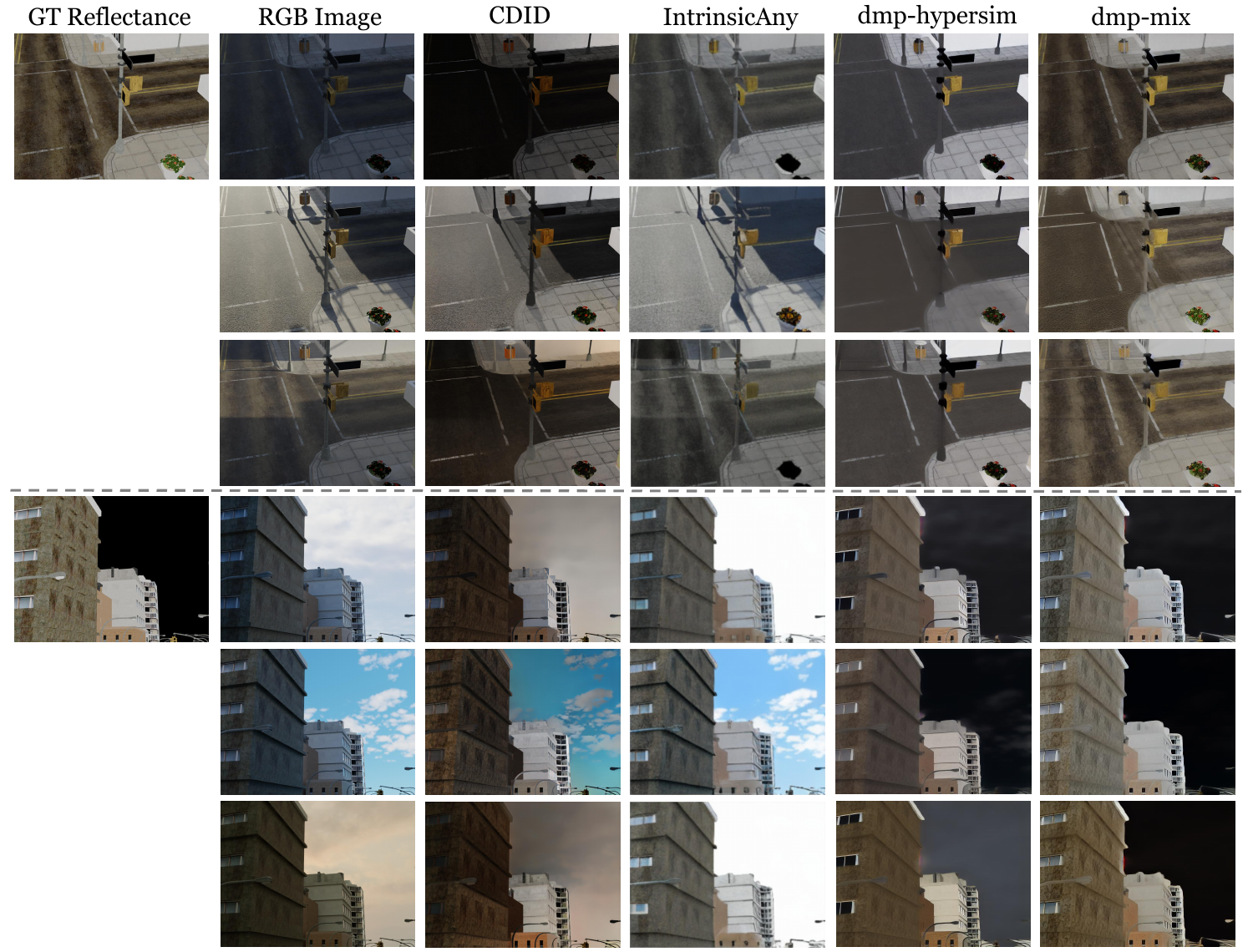}
\caption{Visualization of decomposed albedo of the same view under multi-illuminations for image intrinsic decomposition.}
\vspace{-0.5cm}
\label{fig:intrinsic}
\end{figure*}
\secspace
\section{Experiments}
\secspace

In this section, we explore the challenges of multi-illumination conditions in intrinsic image decomposition, inverse rendering, and urban scene reconstruction.
We adapt SOTA methods to \codename and establish a benchmark for these tasks under multi-illumination constraints, offering insights into their performance under diverse lighting conditions. Baseline details are in the supplementary.

\ssecspace
\subsection{Intrinsic Image Decomposition}
\ssecspace

\mypara{Implementation Details} 
Intrinsic image decomposition of outdoor scenes has been rarely explored, mainly due to the lack of outdoor intrinsic datasets. To investigate the challenges of intrinsic decomposition in outdoor environments and assess existing methods, we conduct two experiments using the \codename reconstruction dataset: mixed-training (train on our dataset) and direct-evaluation (direct test on our dataset). We use DMP~\cite{lee2024exploiting}, a diffusion-based model, and DPF~\cite{chen2023dpf}, a conventional DNN-based approach, as backbones for supervised intrinsic decomposition training on both indoor-only and mixed indoor-outdoor datasets. Besides, to comprehensively evaluate the intrinsic decomposition quality of existing well-pre-trained models in outdoor urban environments, we benchmark three state-of-the-art (SOTA) methods: IntrinsicAny~\cite{chen2024intrinsicanything} (diffusion-based), CDID~\cite{careagaColorful} (DNN-based), and PIE-Net~\cite{das2022pie} (traditional self-supervised method). These methods are assessed across diverse indoor and outdoor datasets, including \codename, to provide a holistic analysis of their effectiveness in real-world intrinsic decomposition.

\mypara{Datasets} For indoor scenes, we use the widely used synthetic Hypersim~\cite{roberts2021hypersim} dataset (simplified as H) as the mixed-training and direct-evaluation dataset, which provides diffuse and shading for various indoor rooms. For outdoor scenes, we use our \codename dataset (L), which provides multi-illumination intrinsics under diverse urban views. Besides, we use IIW~\cite{bell2014intrinsic} (I) and EDEN~\cite{le2021eden} (E) as indoor and outdoor out-of-domain test datasets, BigTime\_v1~\cite{BigTimeLi18} and Waymo Open~\cite{Sun_2020_CVPR} for real-world evaluation.

\mypara{Metrics} For albedo and shading, we use the scale-invariant PSNR (si-PSNR), SSIM, and LPIPS~\cite{zhang2018unreasonable} to estimate their decomposed quality. We also use the scale-invariant MSE (si-MSE) and the scale-invariant LMSE (si-LMSE) to estimate the statistical value error of each component. Besides, to align with previous studies, we also report WHDR as one of the metrics on IIW in the Suppl.~Sec.~4.

\mypara{Intrinsic Evaluation} As shown in Tab.~\ref{tab: lightcity_intrinsic}, DMP~\cite{lee2024exploiting} mixed-finetuned with \codename acheives the best performance for intrinsic decomposition on \codename. In contrast, both the diffusion-based IntrinsicAny~\cite{chen2024intrinsicanything} and the DNN-based methods struggle with urban scene decomposition, with si-PSNRs below 20 for both intrinsics. For EDEN (Tab.~\ref{tab: eden_intrinsic}), the DNN-based intrinsic model achieves the best average performance for albedo decomposition, attributed to the EDEN included in its training datasets,  while DPF~\cite{chen2023dpf} mixed-finetuned with \codename demonstrates superior accuracy in shading decomposition. Notably, despite not being pre-trained on EDEN, DPF and DMP perform the best in shading decomposition. This advantage benefits from \codename's single-view, multi-illumination nature, which provides diverse shading variations in urban scenes and enables the models to learn strong priors for complex lighting effects.
Additional results on the Hypersim indoor dataset in Suppl.~Sec.~4 show that using the DMP backbone, combined with the mixed \codename dataset, improves even indoor intrinsic decomposition. This further shows that learning diverse lighting conditions from a fixed viewpoint strengthens the model’s decomposition ability.
To further investigate the impact of multi-illumination on intrinsic image predictions, we visualize the albedo estimated by different methods under the same viewpoint but with varying illumination conditions as shown in Fig.~\ref{fig:intrinsic}. For images with pronounced lighting effects, both CDID~\cite{careagaColorful} and IntrinsicAny~\cite{chen2024intrinsicanything} struggle to fully disentangle illumination from the predicted albedo, leaving shadows and shading on the street as in the 3rd and 4th columns of Fig.~\ref{fig:intrinsic}. Additionally, given the same scene under different lighting conditions, the albedo predicted by these methods exhibits significant inconsistency, which is likely due to the single-image supervision paradigm commonly adopted in intrinsic decomposition models. In contrast, when mixed-trained with the \codename dataset, DMP produces more consistent albedo predictions across different lighting conditions. This highlights the importance of incorporating multi-illumination outdoor datasets to improve the generalization of conventional intrinsic decomposition methods in unconstrained environments. More decomposition results on BigTime\_v1 and Waymo Open are in Suppl.~Sec.~4.

\subsection{Multi-view Inverse Rendering}
\ssecspace
  
\mypara{Implementation Details} To highlight the challenges of the \codename dataset in outdoor inverse rendering, we evaluate two SOTA multi-view intrinsic-based inverse rendering methods, i.e., NeRF-OSR~\cite{rudnev2022nerf} and GS-IR~\cite{liang2024gs}, on decomposed intrinsic and material attributes.

\mypara{Datasets} We evaluate inverse rendering on four different blocks, F2, F3, E1, and E2, under uniform circular views. The test views are uniformly sampled one out of every 8 images. Since NeRF-OSR requires multi-illumination inputs and GS-IR requires single-illumination inputs, we use identical training viewpoints but render them under multi- and single-illumination for NeRF-OSR and GS-IR, respectively. For fair evaluation, the same novel fixed-illumination test views are used for both methods.

\mypara{Metrics} We evaluate rendering quality using three metrics: Peak Signal-to-Noise Ratio (PSNR), Structural Similarity (SSIM), and LPIPS~\cite{zhang2018unreasonable}. We also assess the quality of intrinsic decomposition for inverse rendering. Specifically, both GS-IR and NeRF-OSR recover albedo as a light-independent intrinsic property. For light-dependent components, NeRF-OSR explicitly models illumination and shadows to account for varying lighting conditions, whereas GS-IR incorporates material properties, i.e., metallic and roughness as key factors in light interaction. We use MSE as the evaluation metric for material properties.

\begin{figure}[t]
\centering
\includegraphics[width=\linewidth]{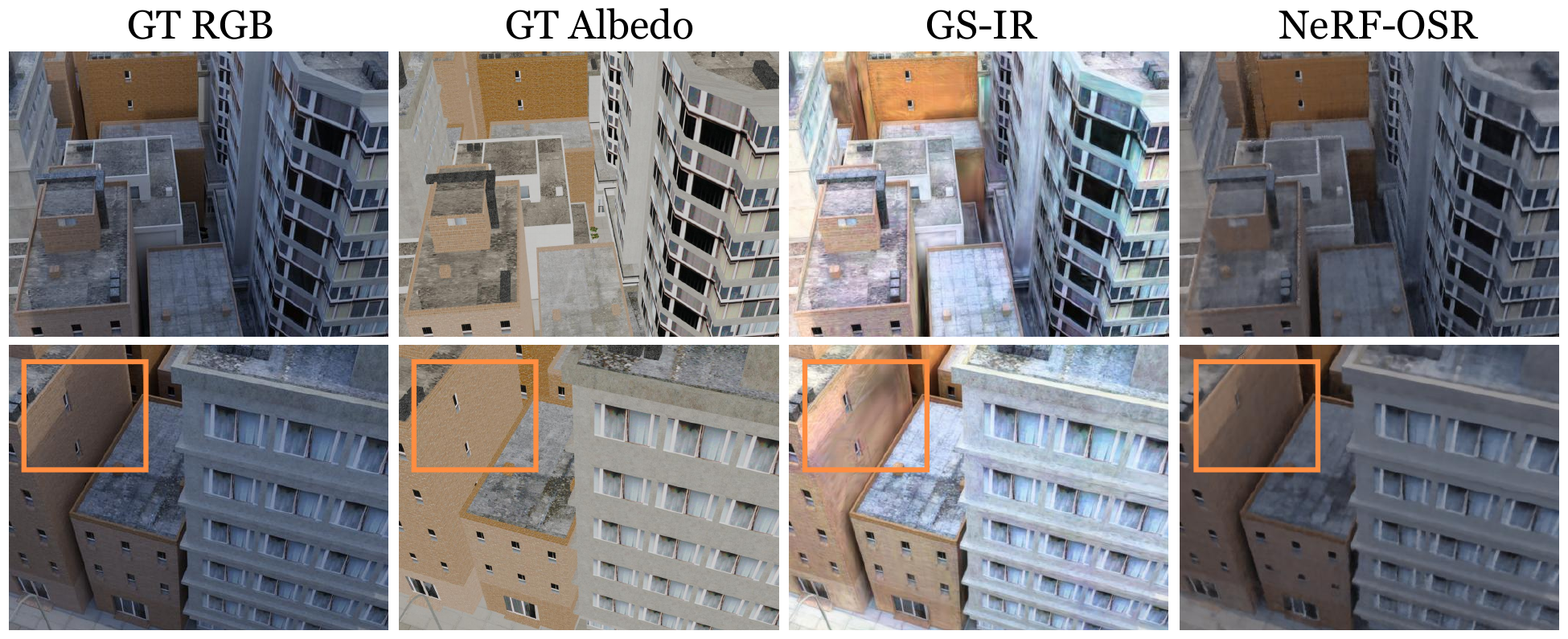}
\caption{Visualization of albedo from inverse rendering.}
\label{fig:pbr_rendering}
\vspace{-2.0em}
\end{figure}

\mypara{Inverse Rendering} To investigate the impact of different illumination disentanglement strategies on intrinsic decomposition, we evaluate the albedo reconstruction quality for both methods, as shown in Tab.~\ref{tab:inverse_pbr11}. Across various scenes, NeRF-OSR achieves a higher average PSNR, while GS-IR demonstrates superior visual quality compared to NeRF-OSR’s image-dependent albedo prediction via neural networks for higher SSIM and LPIPS.
However, as shown in Fig.~\ref{fig:pbr_rendering}, during multi-view optimization in urban scenes, both methods struggle with physically-based, scene-specific intrinsic decomposition, often entangling illumination effects with the diffuse color.
Additionally, GS-IR’s Gaussian-based volumetric rendering enables the synthesis of more realistic images; however, it remains susceptible to light leakage.
In contrast, NeRF-OSR’s image-based disentanglement approach tends to introduce surface roughness and noise in the rendered results, primarily due to insufficient separation between shadows and albedo. Besides, we observe that GS-IR achieves superior performance in both novel view synthesis and geometric reconstruction compared to NeRF-OSR across all four blocks. 
Detailed results on material estimation, geometry reconstruction, and novel view synthesis are provided in Suppl.~Sec.~5.

\begin{table}
\centering
\small
\resizebox{0.7\linewidth}{!}{
\tabcolsep 3pt
\begin{tabular}{c|cc|cc} 
\toprule
Block    & \multicolumn{2}{c|}{F2}         & \multicolumn{2}{c}{F3}           \\ 
\hline
Methods & NeRF-OSR~\cite{rudnev2022nerf}       & GS-IR~\cite{liang2024gs}          & NeRF-OSR       & GS-IR           \\ 
\hline
PSNR    & 15.11          & \textbf{15.32} & \textbf{15.34} & 13.77           \\
SSIM     & 0.583          & \textbf{0.630} & 0.641          & \textbf{0.700}  \\
LPIPS    & 0.362          & \textbf{0.312} & 0.343          & \textbf{0.275}  \\ 
\hline
Block    & \multicolumn{2}{c|}{E1}         & \multicolumn{2}{c}{E2}           \\ 
\hline
Methods & NeRF-OSR       & GS-IR          & NeRF-OSR       & GS-IR           \\ 
\hline
PSNR     & \textbf{15.00} & 12.85          & \textbf{13.92} & 13.10           \\
SSIM    & 0.595          & \textbf{0.631} & 0.555          & \textbf{0.625}  \\
LPIPS    & 0.385          & \textbf{0.340} & 0.430          & \textbf{0.349}  \\
\bottomrule
\end{tabular}
}
\caption{Comparisons of albedo estimation on \codename}
\label{tab:inverse_pbr11}
\vspace{-1.5em}
\end{table}


\ssecspace
\subsection{Multi-illumination Urban Reconstruction}
\ssecspace

\mypara{Implementation Details} 
To reveal challenges of \codename dataset in outdoor reconstruction under multi-illumination, we evaluate five SOTA reconstruction models, i.e. NeRF-W~\cite{martin2021nerf}, NeRF-OSR~\cite{rudnev2022nerf}, Wild-gaussian~\cite{kulhanek2024wildgaussians}, Gaussian-wild~\cite{zhang2024GS-W}, NexusSplats~\cite{tang2024nexussplats}, and analyze their performance on novel view synthesis and geometry recovery. 

\mypara{Datasets} To minimize the impact of highly challenging scenes on reconstruction quality and better assess each method's ability to handle multi-illumination, we select four relatively small blocks in our \codename dataset, i.e. F2, F3, E1, and E2, and generate uniform circle camera views for reconstruction. 
Each block consists of around 200 images with every 8th image chosen as a test view. 
To address the gap in evaluating novel view synthesis quality for urban reconstruction under natural lighting variations, we construct two groups of data for each block under the same sample views: single-illumination and multi-illumination.
For the multi-illumination setting, each view randomly selects an environment map from our collection.
In contrast, the single-illumination setting uses a single, well-lit sunny HDRI map, distinct from those in the multi-illumination setting, to ensure a controlled and unambiguous lighting condition.
We train all methods under two illumination conditions and the training parameters strictly follow the settings of the original paper.

\mypara{Metrics} For novel view synthesis, we use PSNR, SSIM, and LPIPS as evaluation metrics, and follow the evaluation setting of NeRF-W.
For geometry reconstruction, we assess the accuracy of normals using two metrics: Mean Angular Error (MeaAE) and Median Angular Error (MedAE). We evaluate all methods under test views from both the single- and multi-illumination settings.

\setlength{\tabcolsep}{3pt}
\begin{table*}[t]
\centering
\small
\vspace{-0.5em}
\resizebox{0.9\linewidth}{!}{
\tabcolsep 3pt
\begin{tabular}{clcccccc|cccccc} 
\hline
\multicolumn{2}{c|}{Block}   & \multicolumn{6}{c|}{F2~}  & \multicolumn{6}{c}{F3}  \\ 
\hline
\multicolumn{2}{c|}{Lighting}      & \multicolumn{3}{c}{Multi Illumination}   & \multicolumn{3}{c|}{Single Illumination}    & \multicolumn{3}{c}{Multi Illumination}    & \multicolumn{3}{c}{Single Illumination}  \\ 
\hline
\multicolumn{2}{c|}{Metrics}       & \multicolumn{1}{l}{PSNR$\uparrow$}        & \multicolumn{1}{l}{SSIM$\uparrow$}        & \multicolumn{1}{l}{LPIPS$\downarrow$}     & \multicolumn{1}{l}{PSNR$\uparrow$}        & \multicolumn{1}{l}{SSIM$\uparrow$}        & \multicolumn{1}{l|}{LPIPS$\downarrow$}    & \multicolumn{1}{l}{PSNR$\uparrow$}        & \multicolumn{1}{l}{SSIM$\uparrow$}        & \multicolumn{1}{l}{LPIPS$\downarrow$}     & \multicolumn{1}{l}{PSNR$\uparrow$}        & \multicolumn{1}{l}{SSIM$\uparrow$}        & \multicolumn{1}{l}{LPIPS$\downarrow$}      \\ 
\hline
\multicolumn{2}{c|}{Gaussian-wild~\cite{zhang2024GS-W}}          & {\cellcolor{tabsecond}}25.72 & {\cellcolor{tabfirst}}0.844 & {\cellcolor{tabsecond}}0.308 & {\cellcolor{tabsecond}}21.54 & {\cellcolor{tabsecond}}0.794 & {\cellcolor{tabsecond}}0.331 & {\cellcolor{tabsecond}}28.62 & {\cellcolor{tabfirst}}0.903 & {\cellcolor{tabfirst}}0.154 & {\cellcolor{tabsecond}}24.88 & {\cellcolor{tabfirst}}0.859 & {\cellcolor{tabfirst}}0.191  \\
\multicolumn{2}{c|}{Wild-gaussian~\cite{kulhanek2024wildgaussians}}       & 22.20                                     & 0.759                                     & 0.483                                     & 18.80                                     & 0.691                                     & 0.545                                     & 25.93                                     & 0.852                                     & 0.312                                     & 22.06 & 0.787 & 0.381  \\
\multicolumn{2}{c|}{NexusSplats~\cite{tang2024nexussplats}}   & 23.70                                     & 0.779                                     & 0.476                                     & 20.59 & 0.713 & 0.517 & 25.06                                     & 0.815                                     & 0.393                                     & 21.43                                     & 0.742                                     & 0.458                                      \\
\multicolumn{2}{c|}{Nerf-W~\cite{martin2021nerf}}        & {\cellcolor{tabfirst}}26.06 & {\cellcolor{tabsecond}}0.832 & {\cellcolor{tabfirst}}0.289 & {\cellcolor{tabfirst}}24.368                  & {\cellcolor{tabfirst}}0.797                   & {\cellcolor{tabfirst}}0.296                               & {\cellcolor{tabfirst}}29.07 & {\cellcolor{tabsecond}}0.874 & {\cellcolor{tabsecond}}0.240 & {\cellcolor{tabfirst}}25.90               & {\cellcolor{tabsecond}}0.819                & {\cellcolor{tabsecond}}0.275  \\ 
\multicolumn{2}{c|}{NeRF-OSR~\cite{rudnev2022nerf}}    & 25.40                                     & 0.778                                     & 0.417                                     & 21.08                                    & 0.697                                 & 0.444                                   & 26.07                                    & 0.802                                  & 0.381                                     & 20.95 & 0.703 & 0.418 \\
\hline
 &                                 & \multicolumn{1}{l}{}                      & \multicolumn{1}{l}{}                      & \multicolumn{1}{l}{}                      & \multicolumn{1}{l}{}                      & \multicolumn{1}{l}{}                      & \multicolumn{1}{l}{}                      & \multicolumn{1}{l}{}                      & \multicolumn{1}{l}{}                      & \multicolumn{1}{l}{}                      & \multicolumn{1}{l}{}                      & \multicolumn{1}{l}{}                      & \multicolumn{1}{l}{}                       \\ 
\hline
\multicolumn{2}{c|}{Block}         & \multicolumn{6}{c|}{E1}           & \multicolumn{6}{c}{E2}    \\ 
\hline
\multicolumn{2}{c|}{Lighting}      & \multicolumn{3}{c}{Multi Illumination} & \multicolumn{3}{c|}{Single Illumination}  & \multicolumn{3}{c}{Multi Illumination}  & \multicolumn{3}{c}{Single Illumination} \\ 
\hline
\multicolumn{2}{c|}{Metrics}       & PSNR$\uparrow$                            & \multicolumn{1}{l}{SSIM$\uparrow$}        & \multicolumn{1}{l}{LPIPS$\downarrow$}     & PSNR$\uparrow$                            & SSIM$\uparrow$                            & LPIPS$\downarrow$                         & PSNR$\uparrow$                            & \multicolumn{1}{l}{SSIM$\uparrow$}        & \multicolumn{1}{l}{LPIPS$\downarrow$}     & PSNR$\uparrow$                            & SSIM$\uparrow$                            & \multicolumn{1}{l}{LPIPS$\downarrow$}      \\ 
\hline
\multicolumn{2}{c|}{Gaussian-wild~\cite{zhang2024GS-W}}          & {\cellcolor{tabfirst}}27.78 & {\cellcolor{tabfirst}}0.855 & {\cellcolor{tabfirst}}0.300 & {\cellcolor{tabfirst}}22.99 & {\cellcolor{tabfirst}}0.778 & {\cellcolor{tabfirst}}0.364 & {\cellcolor{tabsecond}}27.94 & {\cellcolor{tabfirst}}0.861 & {\cellcolor{tabfirst}}0.250 & {\cellcolor{tabsecond}}23.09 & {\cellcolor{tabfirst}}0.803 & {\cellcolor{tabfirst}}0.297  \\
\multicolumn{2}{c|}{Wild-gaussian~\cite{kulhanek2024wildgaussians}}       & 24.14                                     & 0.763                                     & 0.521                                     & 19.52                                     & {\cellcolor{tabsecond}}0.655 & 0.619                                     & 24.23                                     & 0.750                                     & 0.550                                     & 19.17                                     & 0.645                                     & 0.641                                      \\
\multicolumn{2}{c|}{NexusSplats~\cite{tang2024nexussplats}}   & 24.91                                     & {\cellcolor{tabsecond}}0.770 & 0.521                                     & 20.13 & 0.653  & 0.605 & 25.36    & 0.787   & 0.462   & 21.00 & 0.694 & 0.541  \\
\multicolumn{2}{c|}{Nerf-W~\cite{martin2021nerf}}       & {\cellcolor{tabsecond}}26.61 & 0.730                                     & {\cellcolor{tabsecond}}0.465 & {\cellcolor{tabsecond}}21.72                                    & 0.599                                     & {\cellcolor{tabsecond}}0.507   & {\cellcolor{tabfirst}}28.71 & {\cellcolor{tabsecond}}0.824 & {\cellcolor{tabsecond}}0.325 & {\cellcolor{tabfirst}}24.12   & {\cellcolor{tabsecond}}0.736   & {\cellcolor{tabsecond}}0.368                                      \\
\multicolumn{2}{c|}{NeRF-OSR~\cite{rudnev2022nerf}}       & 25.88                                    & 0.756                                    & 0.476                                    & 18.85                                   & 0.601                                   & 0.525                                   & 26.07                                   & 00.764                                   & 0.473                                   & 13.53 & 0.517 & 0.558  \\
\hline
\end{tabular}
}
\caption{Comparisons of novel view synthesis under multi-illumination in urban scene. The \colorbox{tabfirst}{first} and \colorbox{tabsecond}{second} values are highlighted.}
\label{tab: multi_illum_novelview}
\vspace{-1.0em}
\end{table*}

\begin{figure*}[th]
\centering
\includegraphics[width=0.95\linewidth]{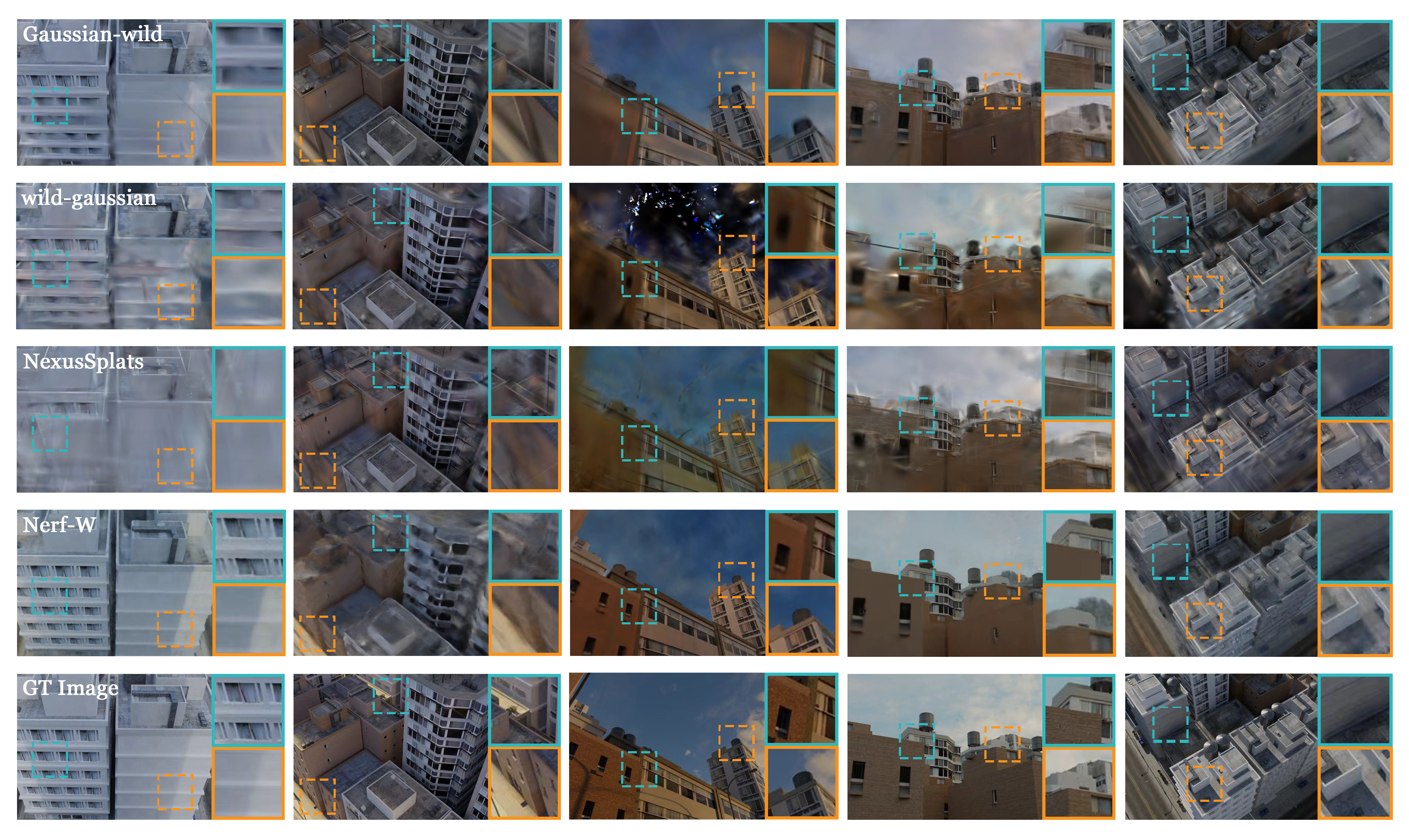}
\vspace{-0.5em}
\caption{Visualization of novel view synthesis under multi-illumination for urban reconstruction.}
\label{fig:multirecon_novelview}
\vspace{-1.5em}
\end{figure*}

\mypara{Novel View Synthesis} We first train all methods under the single-illumination setting as a baseline to assess their fundamental reconstruction quality. 
Our primary focus lies in evaluating the rendering quality of different methods trained on the multi-illumination dataset. 
As shown in Tab.~\ref{tab: multi_illum_novelview}, we assess novel view synthesis performance using two test sets: (1) multi-illumination test set, which follows the same lighting distribution as the training set, and (2) single-illumination test set, which introduces a simple yet challenging lighting condition for controlled evaluation. 

Horizontally, all methods exhibit consistently higher PSNR on the multi-illumination test set, compared to the single-illumination test set. 
It suggests that current SOTA multi-illumination reconstruction methods struggle to generalize to unseen lighting conditions.
This limitation likely arises from their reliance on image-dependent appearance priors learned from the training set, which is inherently distribution-specific and fails to adapt effectively when exposed to different illumination conditions.

Vertically, on relatively smaller-scale blocks F2 and F3, NeRF-W~\cite{martin2021nerf} achieves higher PSNR than GS-based methods, indicating its strong performance in simpler scenes with lower geometry complexity.
However, on more intricate urban blocks E1 and E2, which feature denser building structures and richer textures, NeRF-W~\cite{martin2021nerf} experiences a slight PSNR drop compared to Gaussian-wild~\cite{zhang2024GS-W}.
Meanwhile, Gaussian-wild~\cite{zhang2024GS-W} demonstrates superior SSIM and LPIPS scores, indicating its ability to better preserve perceptual and structural fidelity of complex scenes. 

In general, as shown in Fig.~\ref{fig:multirecon_novelview}, NeRF-W effectively captures the global lighting effects (e.g., shadows cast on buildings) (col.~1\&2) and remains robust to simpler scenes (e.g., fine-detail reconstruction) (col.~3\&4). In contrast, Gaussian-wild adapts better to complex scenes at the cost of increased sensitivity to multi-view inconsistencies. Please see Suppl.~Sec.~6 for geometry evaluation and baselines.

\secspace
\section{Conclusion}
\secspace

In this paper, we propose \codename, a high-quality urban scene dataset with rich lighting variations. It surpasses previous datasets in terms of both illumination complexity and attribute diversity. Leveraging these rich attributes, we evaluate and benchmark three key tasks under complex urban lighting: intrinsic decomposition, inverse rendering, and outdoor reconstruction. Our findings highlight the challenges that existing methods face in maintaining result consistency under multi-illumination. 
Experiments show that our dataset enhances the intrinsic decomposition quality and supports both NeRF and 3DGS for inverse rendering and reconstruction evaluation in urban scenes with variable lighting.
We hope that future research will focus on improving the robustness of core tasks in urban scenes with complex lighting variations.

\paragraph{Acknowledgement} This work was partially supported by the NSF of China (No.~62425209 and No.~62441222), Information Technology Center and State Key Lab of CAD\&CG, Zhejiang University.

{

 \small \bibliographystyle{ieeenat_fullname} \bibliography{main}
}

\setcounter{section}{0}

\section{More for Related Work}

\subsection{Intrinsic Decomposition}

Intrinsic decomposition aims to separate an image into reflectance (albedo), shading and sometimes additional components. Traditional intrinsic decomposition methods rely on different assumptions, leading to three main models: grayscale intrinsic models, RGB intrinsic models, and residual models.
Grayscale intrinsic models were widely used in early works, with optimization-based approaches such as ~\cite{bell2014intrinsic} and various data-driven methods~\cite{li2018cgintrinsics} estimating reflectance and shading under a single-channel assumption. RGB intrinsic models address the limitations of grayscale models by explicitly estimating diffuse color and shading variations, leading to improved accuracy in non-uniform lighting conditions. However, both grayscale and RGB models rely on the Lambertian assumption, making them inadequate for handling specular reflections. To overcome this, residual intrinsic models~\cite{shi2017learning, yeh2022learning, careagaColorful},  were introduced, decomposing an image into albedo 
$A$, shading $S$, and a residual term $R$ to better account for specular effects. Several works have explored this decomposition for improved reflectance modeling.
Despite advancements, most intrinsic decomposition and inverse rendering approaches are evaluated on simple indoor datasets due to data limitations. Expanding these methods to complex outdoor scenes remains an ongoing challenge, particularly under diverse illumination conditions.

\subsection{Inverse Rendering}

Inverse rendering aims to recover intrinsic scene properties such as albedo, shading, and material properties from images, enabling applications like relighting and novel view synthesis. While significant progress has been made, most existing methods and evaluations remain object-centric, with limited exploration in large-scale, complex outdoor environments.

Early works in inverse rendering relied on physics-based models and optimization techniques to estimate reflectance and shading from single images~\cite{barron2014shape, lombardi2018deep}. With the rise of neural representations, NeRF-based approaches have been developed to jointly learn scene geometry and appearance under varying lighting conditions. NeRV~\cite{srinivasan2021nerv} and NeRD~\cite{boss2021nerd} incorporated reflectance decomposition into NeRF, but their evaluations were limited to controlled, object-centric datasets. More recent works, such as PhySG~\cite{zhang2021physg}and InvRender~\cite{zhang2022modeling}, extended inverse rendering to handle non-Lambertian surfaces and indirect illumination, yet their experiments remained focused on synthetic or small-scale real-world objects.

Gaussian-based representations have also been explored for inverse rendering. GS-IR~\cite{liang2024gs} and Relit3DGS~\cite{gao2024relightable}extended 3D Gaussian Splatting for relighting by decomposing scene appearance into intrinsic components. However, these methods are still constrained to object-level reconstructions and have not been tested on large-scale outdoor environments.

Despite these advancements, inverse rendering has yet to be widely explored in large, real-world scenes. Existing datasets are predominantly object-centric (e.g., DTU~\cite{jensen2014large}, NeRF Synthetic~\cite{mildenhall2021nerf}, OmniObject3D~\cite{wu2023omniobject3d} , limiting the generalization of these methods to urban-scale outdoor environments. The lack of benchmarks with complex outdoor lighting and diverse materials remains a significant barrier to extending inverse rendering beyond object-level scenes.

\subsection{Outdoor Scene Reconstruction}

Outdoor scene reconstruction has been widely studied, with Neural Radiance Fields (NeRF)~\cite{mildenhall2021nerf} and 3D Gaussian Splatting (3DGS)~\cite{kerbl20233d} enabling high-quality scene representation. 
Methods like CityNeRF~\cite{turki2022mega} and CityGaussian~\cite{liu2024citygaussian} further enhance large-scale urban reconstruction. 
However, real-world urban-scale data collection inherently involves complex lighting variations due to weather, time of day, and environmental factors. The presence of inconsistent illumination poses significant challenges for outdoor scene reconstruction.
To address illumination variations, recent works integrate appearance modeling. NeRF-W~\cite{martin2021nerf} first introduced latent embeddings for variational lighting appearance. 
Ha-NeRF~\cite{chen2022hallucinated}, CR-NeRF~\cite{yang2023cross} an K-Planes~\cite{fridovich2023k} leveraged CNN-based, cross-ray paradigm, and feature grids to modeling different lighting effects respectively. 
NeuralRecon~\cite{sun2022neural} focused on geometry reconstruction under uncontrolled conditions. 
More recently, efforts to extend 3DGS with appearance modeling have merged, including wild-gaussians~\cite{xu2024wild}, Wild-GS~\cite{kulhanek2024wildgaussians}, Gaussian-wild~\cite{zhang2024GS-W}, and SWAG~\cite{dahmani2024swag}. 
While these methods improve robustness on datasets like Phototourism~\cite{snavely2006photo}, reconstructing urban scenes under extreme multi-illumination conditions remains challenging due to the lack of standardized datasets and uniform benchmarks.

\section{Details for Camera Generation}

\begin{figure*}[t]
\centering
\tiny
\includegraphics[width=\linewidth]{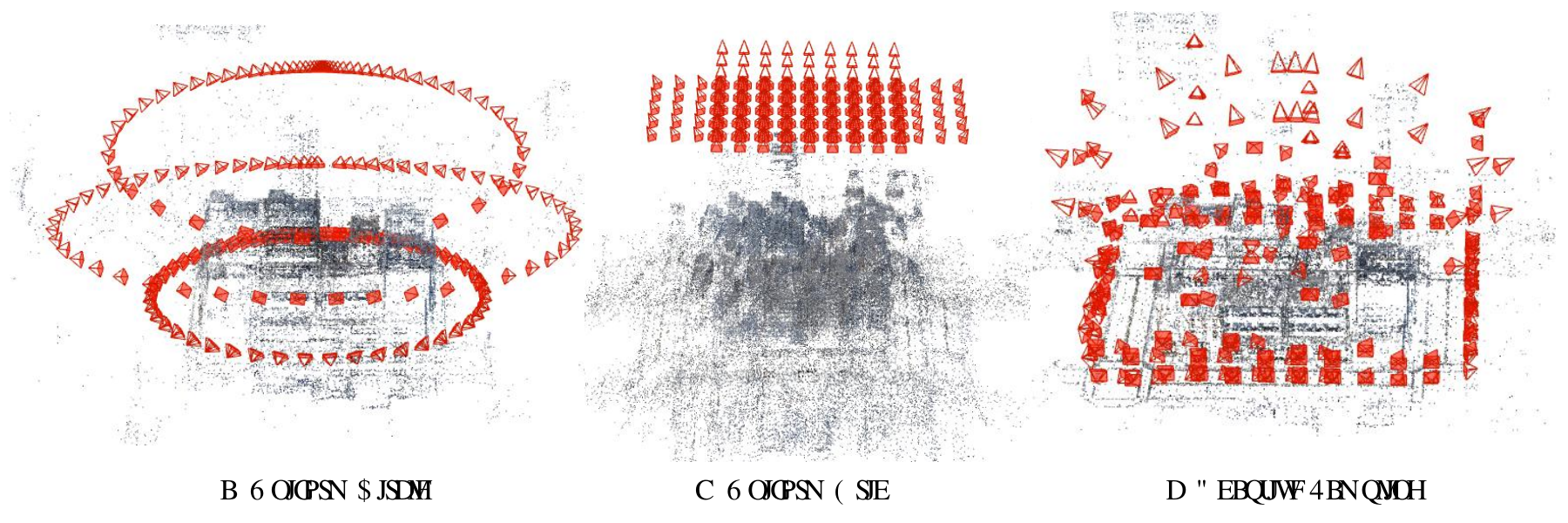}
\caption{The COLMAP coarse point clouds of block F2 under our three types of camera views sampling methods. From left to right represents uniform circle, uniform grid, and adaptive sampling. Our adaptive sampling has the most detailed and uniform point clouds, while the other two cluster on top part of the target scene. }
\label{fig: view_sampling}
\end{figure*}

To generate camera views for our urban scenes, we design two types of view sampling methods, namely uniform view sampling and adaptive sampling. And we display the coarse point cloud reconstructed by COLMAP given our camera intrinsics and extrinsics, as shown in Fig.~\ref{fig: view_sampling}.

\subsection{Uniform View Sampling}
For circular views, we apply two tracking constraints to the cameras and use a frame queue to record their poses. 
The first constraint is based on a Bezier circle path for tracking. 
We place 3 Bezier circles at the center of different regions, with their radius set according to the length and width of the block. 
The heights of the Bezier circles are determined by the maximum object height in the region, ensuring comprehensive views from both a top-down and bottom-up perspective. 
The second constraint is based on the standard object tracking. 
We place an empty object at the center of the scene, allowing the camera to maintain the correct pose while following the Bezier curve. 
The view density of the curve is set adaptively based on the scale of the block.
For grid views, we compute the 2D bounding box of each block and divide this bounding box into grids of varyingg resolutions based on its scale hierarchy. 
Within each grid, we place four cameras, with pitch angles ranging from 20 to 45 degrees and yaw angles of [0, 90, 180, 270] degree, respectively.

\subsection{Adaptive View Sampling}
For street views, cameras are placed along the streets within each block at 0.5m intervals. 
To enhance the details of the streets and surrounding buildings, we randomly generate cameras oriented in four directions, with heights sampled within the ranges of [0.5m, 0.6m] and [0.9m, 1.3m], and pitch angles in the range of [45, 60] degree. 
For aerial views, note that the uniform view sampling described in the previous paragraph struggles to fully capture the complex occlusion relationships within densely clustered buildings. 
Motivated by this, we aim to adaptively position cameras within densely clustered buildings. 
Specifically, we construct an adjacency lookup table in recursion based on the heights and relative positions of all buildings within a block. 
This lookup table enables us to generate a simplified spatial representation of the block and efficiently identify adjacent structures in four directions. 
For buildings located next to streets, we sample street-facing views at an adjustable height above the building. 
For buildings positioned adjacent to one another, we generate camera poses based on relative height relationships, ensuring finer-grained coverage of intra-block architectural structures. 

\section{Details for LightCity}

\begin{figure}[t]
\centering
\includegraphics[width=0.5\linewidth]{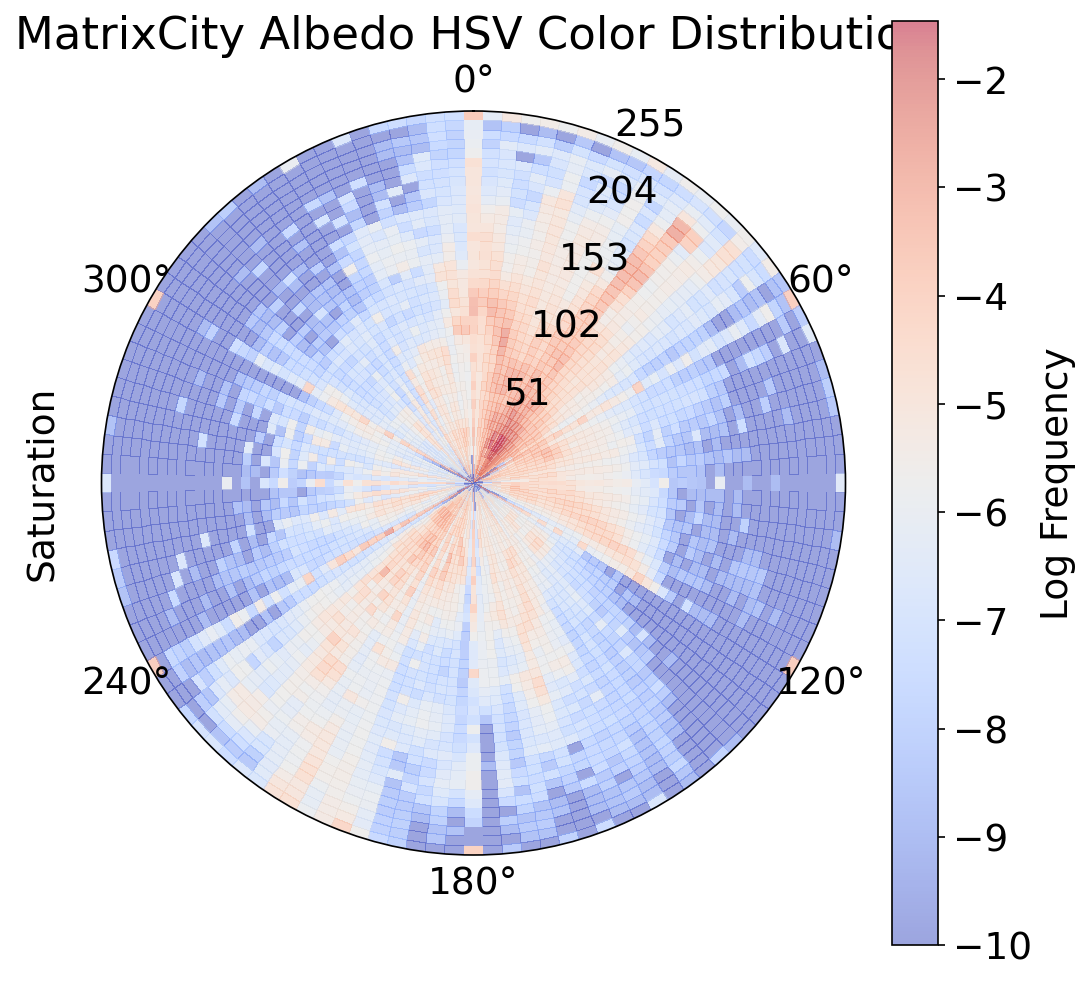}
\caption{Visualization of HSV distribution of MatrixCity albedo images.}
\label{fig:matrixcity_hsv}
\end{figure}

\begin{figure}[t]
\centering
\tiny
\includegraphics[width=0.5\linewidth]{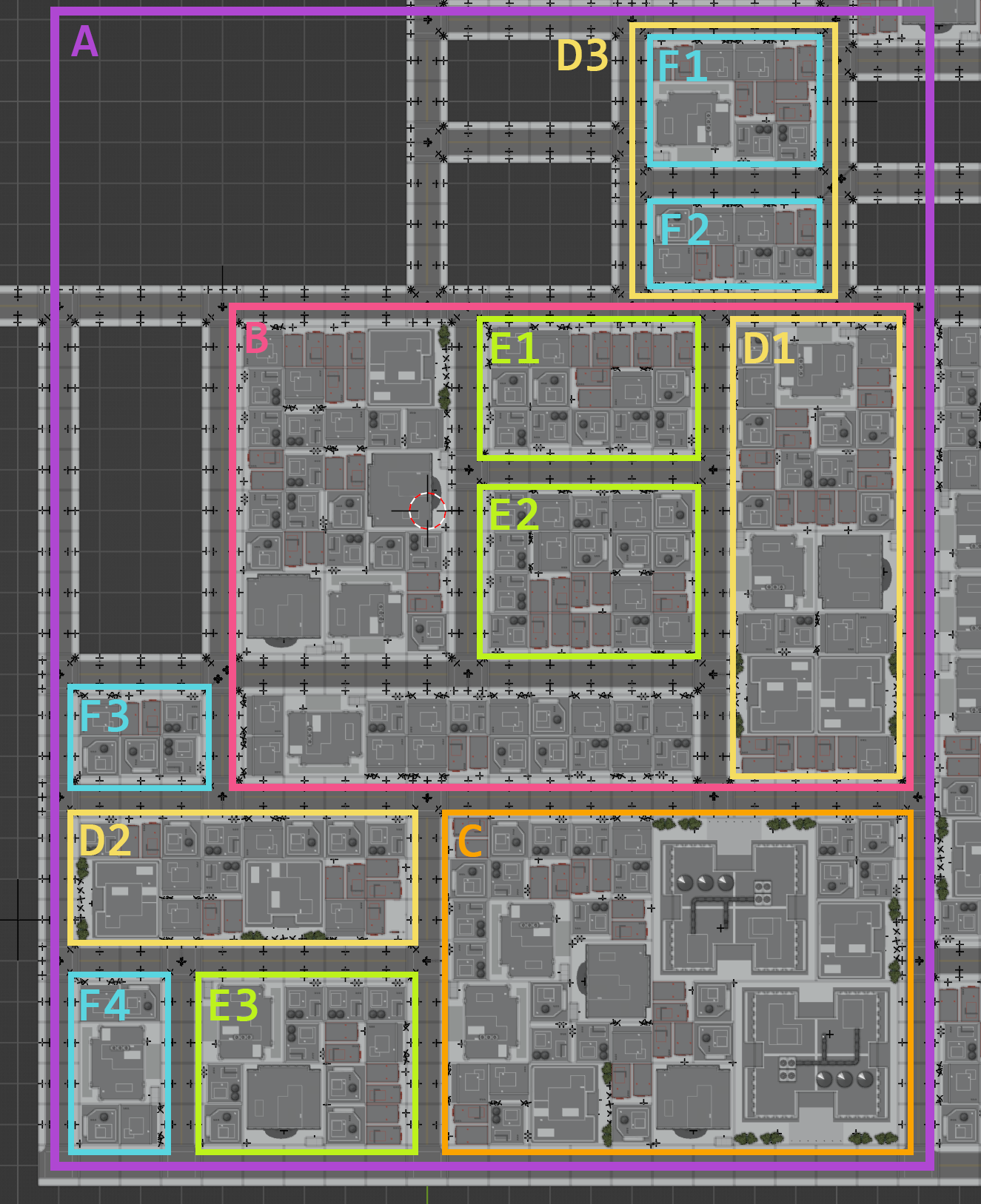}
\caption{Different hierarchies of our \codename reconstruction dataset, divided by different clusters of the scene. The purple rectangle represents a father node, block $A$, of the dataset, which contains images of the whole scene. According to different block size, block $A$ is further split into 5 hierarchies of different scales, namely $B, C, D, E, F$. In total, we have 13 blocks. And we perform urban scene reconstruction on the second smallest $E, F$ hierarchies. }
\label{fig: block_division}
\end{figure}

Our \codename dataset contains two parts, namely the \codename reconstruction dataset and the \codename intrinsic dataset. The dataset for urban scene reconstruction divided into regions based on scene clusters, as shown in Fig.~\ref{fig: block_division}.
To further illustrate our dataset's diverse diffuse color, we also visualize HSV of MatrixCity dataset in Fig.~\ref{fig:matrixcity_hsv}.

\subsection{LightCity Reconstruction Dataset}

The \codename reconstruction dataset is mainly established for task of urban scene reconstruction under multi-illuminations. Under the hierarchical-division of the city assets, we render multi-view images by uniform circle, uniform grid and adaptive sampling. Under the same viewpoints, we also construct a dataset under single-illumination.

\subsection{LightCity Intrinsic Dataset}

The \codename intrinsic dataset is collected for enhancing and benchmarking outdoor intrinsic image decomposition task. To emphasize the challenge of multi-illuminations introduced in the prediction of albedo and shading, we randomly choose two sky environments, randomly rotate each fourth, randomly set the ambient lighting intensity for each view. This type of strategy enables use to simulate the complex lighting interactions within the scene across a day. For each view, we have 8 different lighted images. 

\subsection{Extension of LightCity Dataset}
\begin{figure*}
    \centering
    \includegraphics[width=\linewidth]{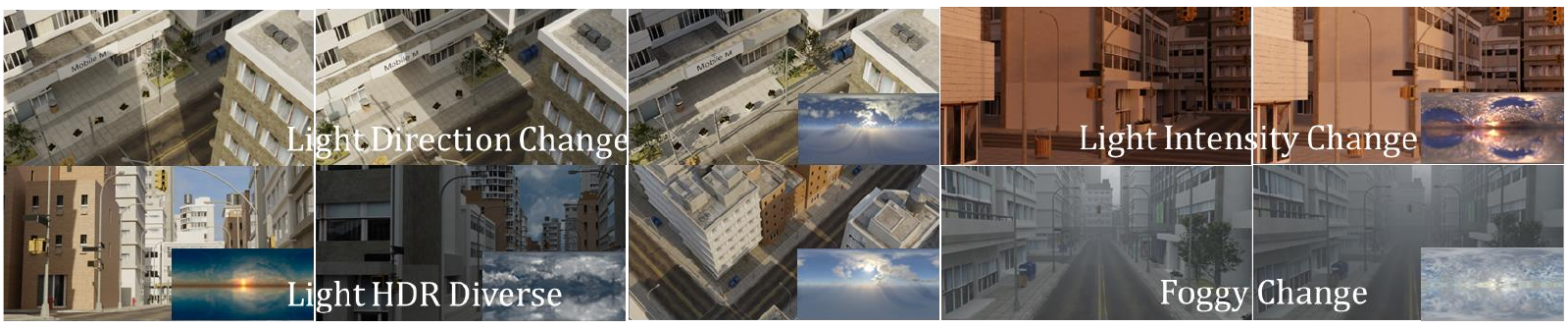}
    \caption{More examples on condition changing of \codename.}
    \label{fig:more_example}
\end{figure*}

\begin{figure*}
    \centering
    \includegraphics[width=\linewidth]{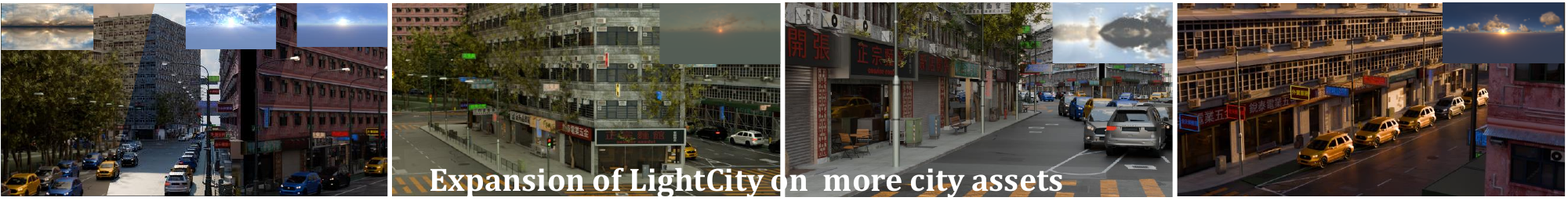}
    \caption{More examples on expansion of \codename built by the City Generator.}
    \label{fig:expand}
\end{figure*}

Since \codename primarily targets the impact of diverse lighting on reconstruction and decomposition, we further include a small subset of renderings under extreme weather conditions such as fog, rain and snow (see Fig.~\ref{fig:more_example}). This enhancement aims to support further studies on weather-aware modeling.
In addition, to enrich scene diversity, we also provide an extension as shown in Fig.~\ref{fig:expand} based on city assets built by the City Generator, another Blender add-on, which covers a broader range of urban layouts.

\section{More Results For Intrinsic Image Decomposition}

\subsection{Baseline Details}

We display a brief summary of methods we used for evaluation of Intrinsic Image Decomposition.

\mypara{DPF}~\cite{chen2023dpf}
DPF (Dense Prediction Fields) is a novel approach for dense prediction tasks using weak point-level supervision. It leverages point-level supervision for dense prediction by predicting values at queried coordinates, inspired by implicit representations. It enables high-resolution outputs and performs well in semantic parsing and intrinsic image decomposition.

\mypara{dmp}~\cite{lee2024exploiting}
DMP leverages pre-trained text-to-image (T2I) diffusion models as priors for dense prediction tasks. It reformulates the diffusion process with interpolations to create a deterministic mapping between input images and predictions. Using low-rank adaptation for fine-tuning, DMP achieves strong generalizability across tasks like 3D property estimation and intrinsic image decomposition.

\mypara{IntrinsicAny}~\cite{chen2024intrinsicanything}
IntrinsicAnything addresses the challenge of recovering object materials from posed images under unknown lighting. Instead of relying solely on differentiable rendering, it introduces a generative material prior using diffusion models for albedo and specular components. This helps resolve ambiguities in inverse rendering. A coarse-to-fine training strategy further enforces multi-view consistency, leading to more accurate material recovery.

\mypara{CDID}~\cite{careaga2023intrinsic}
CDID tackles intrinsic image decomposition by separating an image into diffuse albedo, colorful diffuse shading, and specular residuals. Unlike prior methods assuming single-color illumination and a Lambertian world, it progressively removes these constraints, enabling more realistic and flexible illumination-aware editing.

\mypara{PIENet}~\cite{das2022pie}
PIE-Net is a deep learning method for detecting feature edges in 3D point clouds by representing them as parametric curves (lines, circles, B-splines). It follows a region proposal approach, first identifying edge and corner points, then ranking them for selection. 

\subsection{Detailed Dataset for Evaluations}
We use multiple indoor and outdoor datasets for a through evaluation on our mixed-finetuning mechanism. And we provide a brief summary of all datasets we used.

\mypara{Hypersim} Hypersim is a large-scale synthetic dataset featuring photorealistic indoor scenes with multi-view RGB images, depth maps, surface normals, and intrinsic decomposition (albedo, shading). It serves as a benchmark for tasks like indoor intrinsic decomposition, depth estimation, and inverse rendering.

\mypara{IIW} IIW is a real-world dataset for intrinsic image decomposition, containing over 5,000 images with human-annotated pairwise reflectance comparisons. It provides a diverse set of unconstrained scenes, making it a key benchmark for evaluating intrinsic decomposition methods.

\mypara{EDEN} EDEN is a multimodal synthetic dataset designed for nature-oriented applications, such as agriculture and gardening. It contains over 300K images from 100+ garden models, annotated with various vision modalities, including semantic segmentation, depth, surface normals, intrinsic colors, and optical flow. The dataset can be used for semantic segmentation and monocular depth prediction.

\subsection{Indoor Scenes}
\label{sec:indoor}
We display the evaluation results of image intrinsic decomposition of indoor scenes of Hypersim and IIW in Tab.~\ref{tab: hypersim_intrinsic} and Tab.~\ref{tab:iiw_intrinsic}, respectively. For Hypersim dataset,  the DNN-based CDID has the best averaged performance on si-PSNR, si-MSE and si-LMSE for albedo decomposition. However, the diffusion-based DMP tends have better visual fidelity with SSIM for albedo higher than 0.53, shading higher than 0.62. It aligns with the high quality of generated images of diffusion models. Besides, the DMP mixfine-tuned with \codename tends to get higher si-PSNR and LPIPS for shading estimation. This findings aligns with previous in outdoor datasets. For IIW dataset, the DMP fine-tuned on Hypersim has the best WHDR score, there is a little quality drop for DMP mixfine-tuned with \codename, we attribute this to the domain gap between the two datasets, which brings chanllenge for diffusion models to learn. However, DPF mixfine-tuned with \codename is 5\% lower on WHDR metrics, exhibiting improved performance.

\begin{table}[th]
\centering
\small
\caption{Performance of alebdo estimation on IIW datasets. The \colorbox{tabfirst}{first}, \colorbox{tabsecond}{second} and \colorbox{tabthird}{third} values are highlighted.}
\label{tab:iiw_intrinsic}
\begin{tabular}{c|c|c|c} 
\toprule
\multicolumn{4}{c}{IIW-Indoor}                                                     \\ 
\hline
\multicolumn{2}{c|}{Method}                             & $D_{train}$ & WHDR / \%  \\ 
\hline
\multirow{4}{*}{DNN Based}       & PIE-Net              & /         & 32.77      \\ 
\cline{2-3}
                                 & \multirow{2}{*}{DPF} & H           & 43.14      \\
                                 &                      & H+L         & 38.502     \\ 
\cline{2-3}
                                 & Intrinsic 2024       &
                                 objects & {\cellcolor{tabthird}}21.33      \\ 
\cline{1-3}
\multirow{3}{*}{Diffusion Based} & \multirow{2}{*}{DMP} & H           & {\cellcolor{tabfirst}} 19.08      \\
                                 &                      & H+L         & {\cellcolor{tabsecond}} 20.37      \\ 
\cline{2-3}
                                 & IntrinsicAnything    & /         & 27.08      \\
\bottomrule
\end{tabular}
\end{table}

\setlength{\tabcolsep}{2pt}
\begin{table*}
\centering
\small
\caption{Single image intrinsic decomposition results under Hypersim Indoor dataset. The \colorbox{tabfirst}{first}, \colorbox{tabsecond}{second} and \colorbox{tabthird}{third} values are highlighted.}
\label{tab: hypersim_intrinsic}
\arrayrulecolor{black}
\begin{tabular}{c|c|c|ccccc|ccccc} 
\hline
\multicolumn{13}{c}{Hypersim-Indoor}   \\ 
\hline
\multicolumn{2}{c|}{\multirow{2}{*}{Method}}      & \multirow{2}{*}{$D_{train}$} & \multicolumn{5}{c|}{Albedo}                                                                                                                                                                           & \multicolumn{5}{c}{Shading}                                                                                                                                                                                 \\ 
\cline{4-13}
\multicolumn{2}{c|}{}                             &                     & si-PSNR$\uparrow$                               & SSIM$\uparrow$                                  & LPIPS$\downarrow$                                 & si-MSE$\downarrow$                                & si-LMSE$\downarrow$                               & si-PSNR$\uparrow$                                   & SSIM$\uparrow$                                   & LPIPS$\downarrow$                                  & si-MSE$\downarrow$                                 & si-LMSE$\downarrow$                                 \\ 
\hline
\multirow{4}{*}{DNN}       & PIE-Net              & Outdoor                 & 12.55                                 & 0.449                                 & 0.479                                 & 0.101                                 & 0.095                                 & 14.81~                                 & 0.513~                                 & {\cellcolor{tabthird}}0.454~ & {\cellcolor{tabsecond}}0.025~ & {\cellcolor{tabsecond}}0.024~  \\ 
\hhline{~--~~~>{\arrayrulecolor{tabsecond}}--~>{\arrayrulecolor{tabthird}}-~~~}
                           & \multirow{2}{*}{DPF} & H                   & 15.43                                 & 0.445                                 & 0.576                                 & {\cellcolor{tabsecond}}0.033 & {\cellcolor{tabsecond}}0.031 & 14.59~                                 & {\cellcolor{tabthird}}0.570~ & 0.531~                                 & 0.070~                                 & 0.063~                                  \\
                           &                      & H+L                 & 13.48                                 & 0.403                                 & 0.599                                 & 0.089                                 & 0.082                                 & {\cellcolor{tabthird}}14.85~ & 0.520~                                 & 0.570~                                 & {\cellcolor{tabthird}}0.037~ & {\cellcolor{tabthird}}0.034~  \\ 
\hhline{~>{\arrayrulecolor{black}}-->{\arrayrulecolor{tabfirst}}->{\arrayrulecolor{tabthird}}-->{\arrayrulecolor{tabfirst}}-->{\arrayrulecolor{tabthird}}-~~>{\arrayrulecolor{tabfirst}}--}
                           & CDID           & E+Indoor.etc                & {\cellcolor{tabfirst}}16.70 & {\cellcolor{tabthird}}0.487 & {\cellcolor{tabthird}}0.372 & {\cellcolor{tabfirst}}0.023 & {\cellcolor{tabfirst}}0.020 & {\cellcolor{tabthird}}14.85~ & 0.190~                                 & 0.515~                                 & {\cellcolor{tabfirst}}0.011~ & {\cellcolor{tabfirst}}0.010~  \\ 
\hhline{>{\arrayrulecolor{black}}--->{\arrayrulecolor{tabthird}}->{\arrayrulecolor{tabfirst}}-->{\arrayrulecolor{tabthird}}-->{\arrayrulecolor{tabsecond}}->{\arrayrulecolor{tabfirst}}->{\arrayrulecolor{tabsecond}}-~~}
\multirow{3}{*}{Diffusion} & \multirow{2}{*}{DMP} & H                   & {\cellcolor{tabthird}}16.48 & {\cellcolor{tabfirst}}0.534 & {\cellcolor{tabfirst}}0.369 & {\cellcolor{tabthird}}0.036 & {\cellcolor{tabthird}}0.034 & {\cellcolor{tabsecond}}15.59~ & {\cellcolor{tabfirst}}0.624~ & {\cellcolor{tabsecond}}0.353~ & 0.046~                                 & 0.043~                                  \\
                           &                      & H+L                 & {\cellcolor{tabsecond}}16.56 & {\cellcolor{tabsecond}}0.531 & {\cellcolor{tabsecond}}0.371 & {\cellcolor{tabsecond}}0.033 & {\cellcolor{tabsecond}}0.031 & {\cellcolor{tabfirst}}15.73~ & {\cellcolor{tabsecond}}0.621~ & {\cellcolor{tabfirst}}0.352~ & 0.043~                                 & 0.040~                                  \\ 
\arrayrulecolor{black}\cline{2-3}
                           & IntriAny             & Objects                 & 12.33                                 & 0.407                                 & 0.510                                 & 0.195                                 & 0.182                                 & \multicolumn{5}{c}{/}                                                                                                                                                                                    \\
\hline
\end{tabular}
\end{table*}

\subsection{Sim-to-real Discussion}
Synthetic data plays a vital role in computer and robotic vision, particularly for tasks like scene understanding and inverse rendering. It allows precise control over lighting, materials, and geometry through engines such as Blender or Unreal. However, low-quality synthetic datasets can suffer from a large sim-to-real gap, negatively impacting generalization to real-world images. 
To mitigate this, we have used the best open-sourced rendering engine, Blender Cycles, for photo-realism. This high realism reduces the domain gap and improves transferability, similar to how datasets like Hypersim~\cite{roberts2021hypersim} leveraged PBR to boost real-world performance.

\begin{figure}[h]
    \centering
    \includegraphics[width=\linewidth]{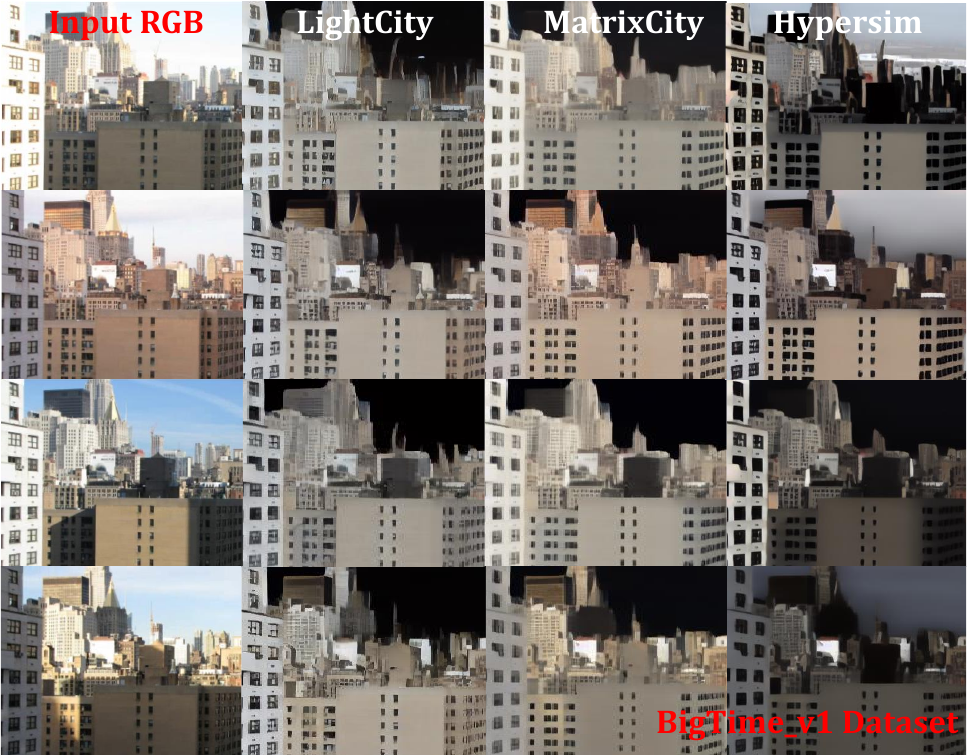}
    \caption{Albedo Decomposed from BigTime\_v1 dataset.}
    \label{fig:intrinsic_bigtime}
\end{figure}

\begin{figure*}[h]
    \centering
    \includegraphics[width=\linewidth]{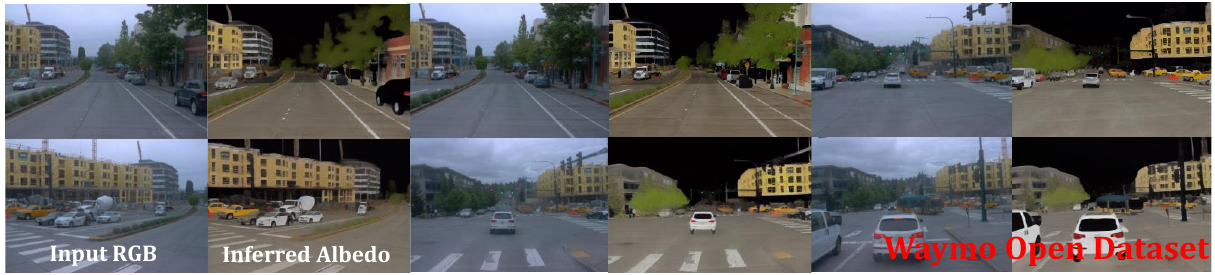}
    \caption{Albedo Decomposed from Waymo Open dataset.}
    \label{fig:intrinsic_waymo}
\end{figure*}

To further evaluate the real-world generalization, we leverage the strong generalization ability of generative models. Recent studies have shown that diffusion models exhibit impressive generalization across domains, including tasks like normal prediction (e.g., StableNormal~\cite{ye2024stablenormal}). 
Building on DMP, a diffusion-based model, we assess intrinsic decomposition performance of both indoor and outdoor real-world scenes. For indoor evaluation, we report results on the IIW dataset(Sec.~\ref{sec:indoor}). For outdoor scenes, we use BigTime\_v1 and the Waymo Open dataset. BigTime\_v1 captures outdoor environments under varying illumination throughout a day, while the Waymo Open dataset offers diverse urban scenes collected under different lighting and weather conditions by Waymo autonomous vehicles.
For albedo consistency as shown in Fig.~\ref{fig:intrinsic_bigtime}, DMP mix-finetuned with \codename presents lowest average variance of 0.015, while mix-finetuned with MatrixCity-mix and purely Hypersim have higher variance of 0.036 and 0.042, respectively.
DMP mix-finetuned with \codename also generalizes well to real-world urban scenes, as shown in Fig.~\ref{fig:intrinsic_waymo}.

Besides, to further minimize the sim-to-real gap, generative models can also be treated as domain transfer models for sim-to-real transfer. Established works have demonstrated the efficacy of such approaches in bridging synthetic-real discrepancies~\cite{zhao2024exploring}. Leveraging diffusion-based image-to-image pipelines such as img2img-turbo~\cite{parmar2024one} and InstructPix2Pix~\cite{brooks2023instructpix2pix} offers a promising future direction to make synthetic datasets more applicable to real-world scenarios.

\section{More Results for Multi-image Inverse Rendering}
\subsection{Baseline Details}

We present a short description for the baseline methods (NeRF-OSR and GS-IR) for our inverse rendering. 

\mypara{NeRF-OSR} is the first approach to learning a neural representation that explicitly decomposes scene geometry, diffuse albedo, and shadows from multi-view and multi-illumination input images, thereby enabling more flexible scene editting.

\mypara{GS-IR} first extends 3DGS for inverse rendering, leveraging a PBR framework to jointly reconstruct scene geometry, material properties, and unknown natural illumination from multi-view captured images at both object-level and scene-level tasks.

\subsection{Novel View Synthesisi and Geometry Quality}
We also provide geometry ground-truth for multi-view inverse rendering. And evaluate the geometry quality of both used baselines. As shown in Tab.~\ref{tab:inverse_novelview}, GS-IR performs better in urban scene inverse rendering than NeRF-based NeRF-OSR both in novel view synthesis and geometry reconstruction.

\subsection{Material Estimation}
As an important component in PBR-based inverse rendering, GS-IR also optimizes per-Gaussian metallic and roughness attribute to produce photo-realistic lighting effect. So we evaluate the decomposed material properties with our ground-truth properties, there are still large step to improve accuracy of material estimation in urban inverse rendering.

\section{More Results for Multi-illumination Outdoor Reconstruction}
\subsection{Baseline Details}

We present a short description for the baseline methods (NeRF-W, wild-gaussians and Gaussian-wild) for our outdoor reconstruction under multi-illumination.                  

\mypara{NeRF-W} extends the implicit NeRF to unconstrained multi-illumination reconstruction by introducing a per-image learned low-dimensional latent appearance embeddings as shared MLP conditions utilizing GLO, thus disentangling scene geometry from illumination inconsistencies.

\mypara{wild-gaussians} adapts the explicit 3D Gaussian Splatting (3DGS) representation for real-world scene reconstruction under varying lighting conditions. It incorporates an MLP-based appearance modeling module with affine color mapping to capture image-dependent Gaussian colors while preserving rendering efficiency.

\mypara{Gaussian-wild} further enhances local high-frequency changes of the scene by separating each Gaussian's appearance into intrinsic and dynamic features based on 3DGS, to better capture fine-grained scene details while adapting to varying lighting conditions.

\mypara{NexusSplats} utilizes an neural network to represent image-specific global lighting conditions and Gaussian-specific localized response to global lighting variations, to effectively capture complex illumination changes across scenes.

\subsection{Novel View Synthesis}

To provide a baseline for our \codename reconstruction dataset. We also train Gaussian-wild (GS-W) under the single-illumination dataset. The result is shown in Tab.~\ref{tab:same_illum}. Compared with that trained under multi-illumination dataset, the performance dropped, which further indicating the strong influence of multi-illumination on performance of urban reconstructions.

\begin{figure*}[t]
\centering
\tiny
\includegraphics[width=\linewidth]{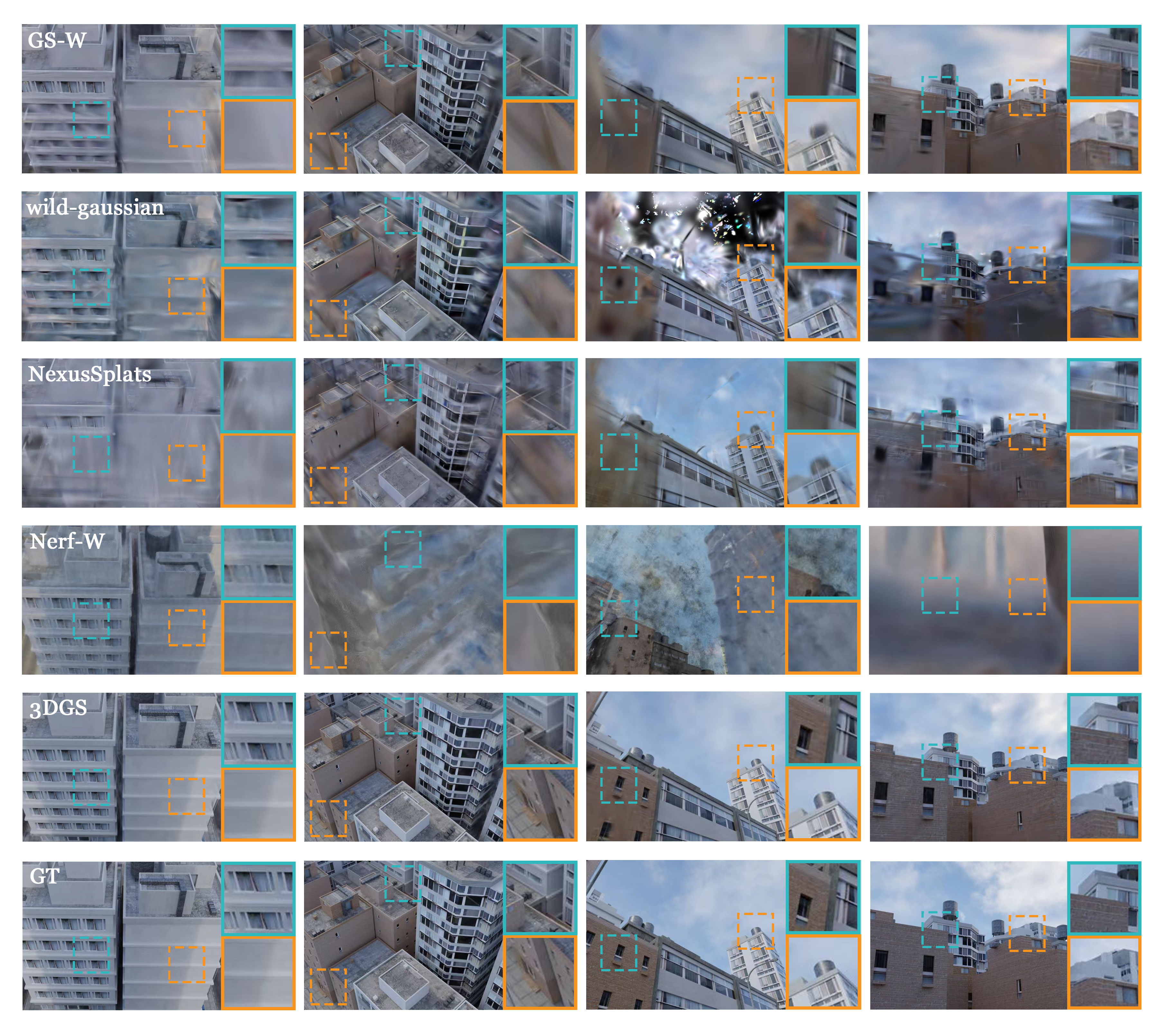}
\caption{Novel-view rendering results under test set of single-illumination dataset.}
\label{fig:same_illum}
\end{figure*}

In previous sections, we display the visualization evaluation results under test set of multi-illumination dataset. We also display the results under the test set of single-illumination dataset in Fig.~\ref{fig:same_illum}. Compared with 3DGS, methods for modeling appearance embedding has a quality degradation. Although, NeRF-W is able to restore the shadow of the image (col1), it's performance under other unseen views remain worse. GS-W tends to restore a more clear structure of the never-seen input GT, but there are floaters in some part. This further illustrate the challenge on our multi-illumination reconstruction dataset.

\begin{figure}[th]
\centering
\includegraphics[width=\linewidth]{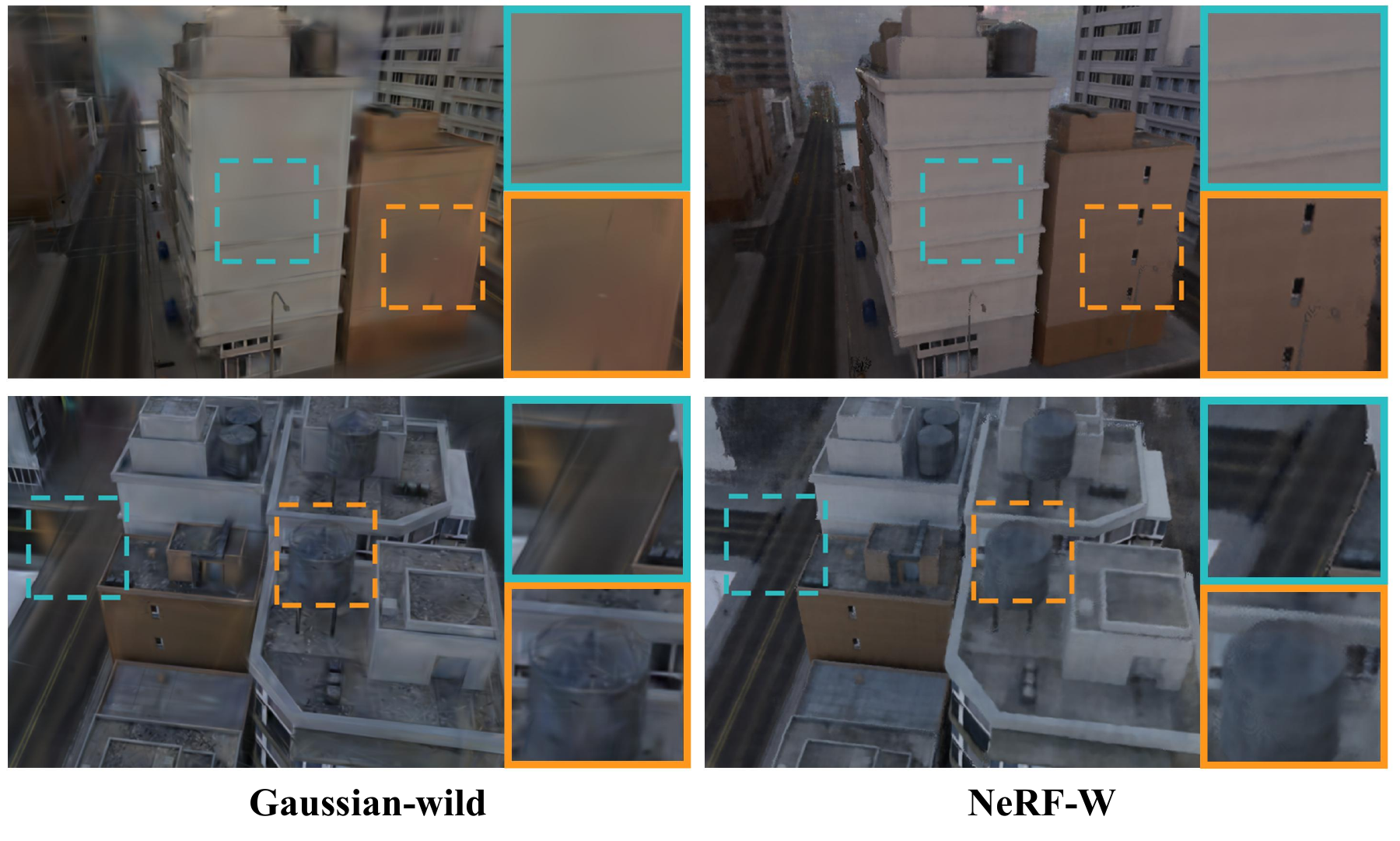}
\caption{Comparison of novel view synthesis under multi-illumination between Gaussian-wild and Nerf-W.}
\label{fig:multirecon_compare}
\end{figure}

We also perform a deep analysis of the performane between the best GS-W and NeRF-W, as visualized in fig.~\ref{fig:multirecon_compare}. The first row illustrates blurred detail of GS-W, floaters covering the building leading to visual artifacts. The second row illustrates blurred detail of NeRF-W, which tends to blur the detail of complex scenes.

\begin{table}[ht]
\centering
\caption{Performance of novel view synthesis of Gaussian-Wild trained under single-illumination dataset for block F2}
\label{tab:same_illum}
\begin{tabular}{l|l} 
\toprule
F2    & Gaussian-wild  \\ 
\hline
PSNR  & 28.74          \\
SSIM  & 0.878          \\
LPIPS & 0.174          \\
\bottomrule
\end{tabular}
\end{table}

\begin{table}
\centering
\caption{Performance of novel view synthesis for multi-view inverse rendering.}
\label{tab:inverse_novelview}
\begin{tabular}{c|c|cccc} 
\toprule
\multicolumn{2}{c|}{}                & \multicolumn{4}{c}{Datasets}                                       \\ 
\hline
Methods                   &   Metrics       & F2~            & F3             & E1             & E2             \\ 
\hline
\multirow{4}{*}{NeRF-OSR} & PSNR     & 17.35          & 20.15          & 21.11          & 20.95           \\
                          & SSIM     & 0.562          & 0.600            & 0.622          & 0.597           \\
                          & LPIPS    & 0.461          & 0.413          & 0.400            & 0.438           \\
                          & MeaAE & 32.81          & 31.39          & 30.96          & 33.98           \\ 
\hline
\multirow{4}{*}{GS-IR}    & PSNR     & \textbf{26.35} & \textbf{27.26} & \textbf{27.29} & \textbf{26.76}  \\
                          & SSIM     & \textbf{0.862} & \textbf{0.90}   & \textbf{0.858} & \textbf{0.861}  \\
                          & LPIPS    & \textbf{0.233} & \textbf{0.186} & \textbf{0.196} & \textbf{0.200}    \\
                          & MeaAE & \textbf{28.07} & \textbf{23.60}  & \textbf{27.73} & \textbf{28.78}  \\
\bottomrule
\end{tabular}
\end{table}

\begin{table}
\centering
\caption{Performance of material estimation for multi-view inverse rendering.}
\begin{tabular}{c|c|c} 
\toprule
Datasets            & Metrics       & \multicolumn{1}{l}{GS-IR}  \\ 
\hline
\multirow{2}{*}{A2} & metallic mse  & 0.1168                     \\
                    & roughness mse & 0.2643                     \\ 
\hline
\multirow{2}{*}{A3} & metallic mse  & 0.1813                     \\
                    & roughness mse & 0.2964                     \\ 
\hline
\multirow{2}{*}{B1} & metallic mse  & 0.2888                     \\
                    & roughness mse & 0.2833                     \\ 
\hline
\multirow{2}{*}{B2} & metallic mse  & 0.2599                     \\
                    & roughness mse & 0.2466                     \\
\bottomrule
\end{tabular}
\end{table}

\setlength{\tabcolsep}{2pt}
\begin{table*}
\centering
\caption{Performance comparison of geometry quality for urban reconstruction under multi-illuminations.}
\label{tab:multi_illum_normal}
\begin{tabular}{c|c|ccccc} 
\toprule
\multicolumn{2}{c|}{}            & \multicolumn{5}{c}{Methods}                                    \\ 
\hline
Datasets            & Metrics    & 3DGS  & Gaussian-wild  & wild-gaussian & NexusSplats & Nerf-w  \\ 
\hline
\multirow{2}{*}{A2} & MeaAE   & 21.57 & \textbf{26.91} & 32.23         & 32.94       & 27.88   \\
                    & MedAE & 21.81 & \textbf{26.38} & 31.30         & 31.81       & 27.49   \\ 
\hline
\multirow{2}{*}{A3} & MeaAE   & 23.50 & \textbf{25.27} & 30.85         & 32.83       & 29.81   \\
                    & MedAE & 23.33 & \textbf{26.00} & 31.08         & 32.22       & 28.45   \\ 
\hline
\multirow{2}{*}{B1} & MeaAE   & 24.34 & \textbf{31.15} & 37.02         & 35.76       & 49.76   \\
                    & MedAE & 23.52 & \textbf{29.65} & 36.31         & 35.45       & 50.55   \\ 
\hline
\multirow{2}{*}{B2} & MeaAE   & 23.99 & \textbf{29.16} & 40.63         & 37.84       & 37.99   \\
                    & MedAE & 23.98 & \textbf{28.13} & 40.34         & 36.23       & 36.21   \\
\bottomrule
\end{tabular}
\end{table*}

\subsection{Geometry Quality}

\begin{figure*}[t]
\centering
\tiny
\includegraphics[width=\linewidth]{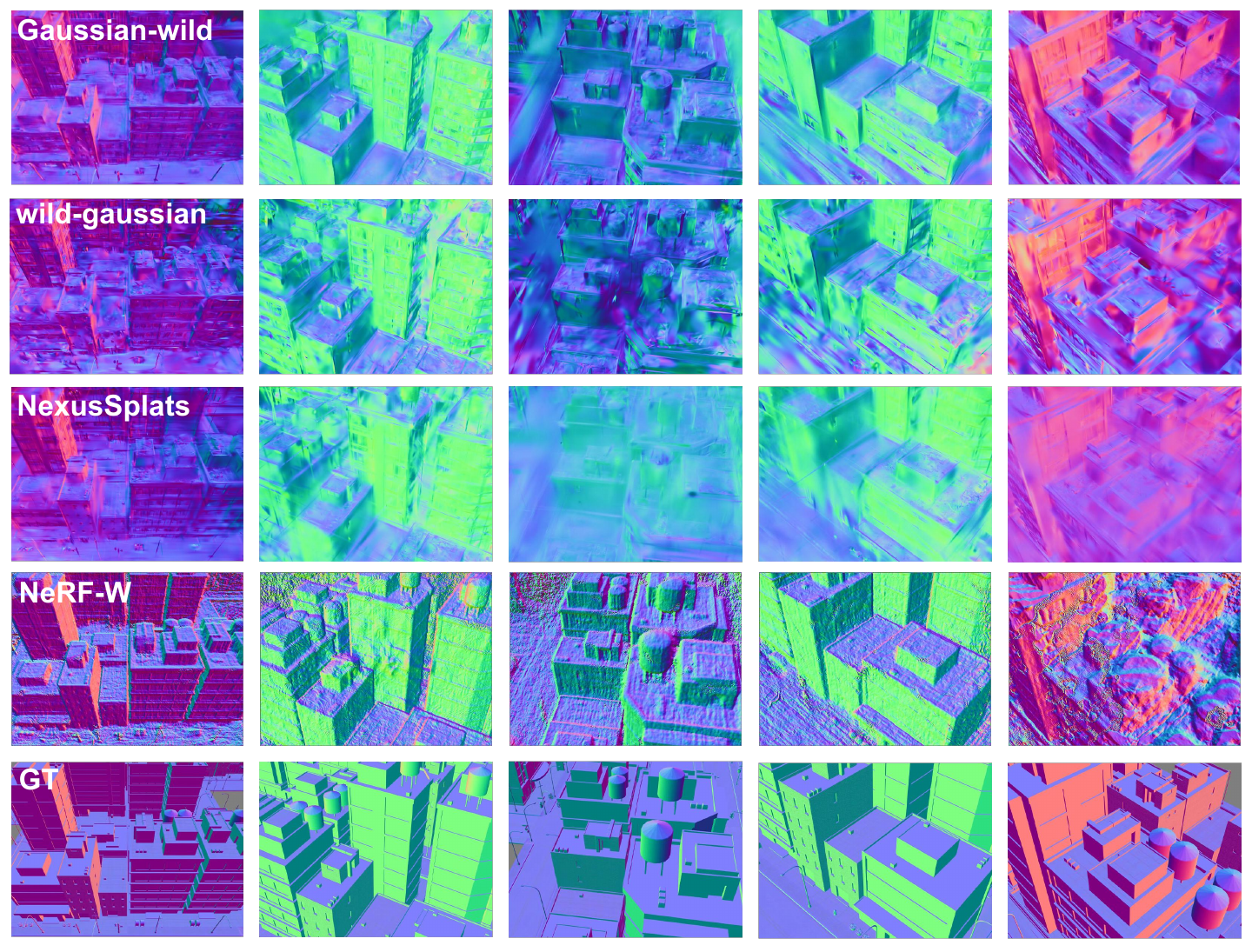}
\caption{Visualization of normal maps for urban reconstruction under multi-illuminations.}
\label{fig:all_normal}
\end{figure*}

For thoroughly evaluate the reconstruction geometry of the \codename dataset, we render the normal map of all used methods under multi-illumination conditions. The error metrics are displayed in Tab.~\ref{tab:multi_illum_normal}. 
Across all blocks, Gaussian-wild  has the lowest MeaAE and MedAE, indicating its prior geometry reconstruction quality compared with other methods. However, under the constraint of multi-illumination, those methods presents a quality decay compared with the origin 3DGS. Besides, we presents the normal map of different methods in fig.~\ref{fig:all_normal}. Although Gaussian-wild has the highest normal accuracy, it tends to be blurred in some flat areas, this may due to the extra floaters introduced by multi-illumination input. However, NeRF-W has a relatively sharp normal except for some roughness. This might be attributed to its discrete sampling of rays. Another two 3DGS-based methods, i.e., wild-gaussian and NexusSplats, can hardly reconstruct normals of the scene, with a wide area of Gaussian surfels covering the screen space (column 3).

\begin{figure*}[h]
    \centering
    \includegraphics[width=\linewidth]{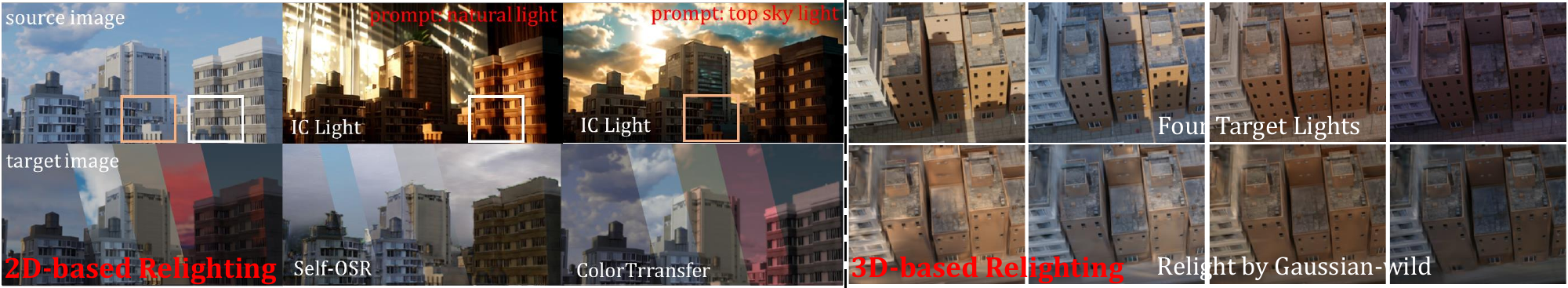}
    \caption{Visual results for 3D-based relighting.}
    \label{fig:relight}
\end{figure*}

\begin{figure}[t]
\centering
\tiny
\includegraphics[width=\linewidth]{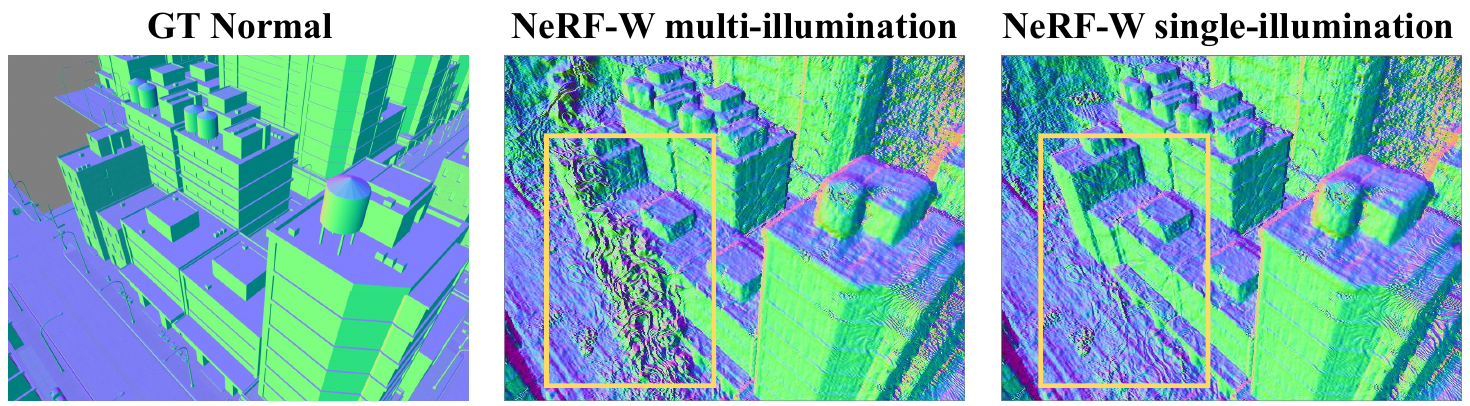}
\caption{Comparisons of normal maps under different views of NeRF-W for urban reconstruction.}
\label{fig:nerf_normal}
\end{figure}

Besides, we also investigate the consistency of normal between different views. As illustrated in Fig.~\ref{fig:nerf_normal}, NeRF-W tends to reconstruct different normal maps between different views of the same scene, exhibiting strong inconsistencies. This problem is not found for GS-based methods since they disentangle apperance and location of each Gaussian.

\section{More Results on Relighting}

Relighting is a vital and real-world task in computer vision, enabliing applications such as content editing, lighting transfer and scene manipulation. In our multi-illumination reconstruction experiments, we perform 3D-based relighting by optimizing the reconstructed representation under test lighting conditions. Visual examples are shown in right of Fig.~\ref{fig:relight}.
In addition to 3D-driven methods like NeRF-OSR, we also evaluate image-based relighting techniques-single-image models that directly manipulate input images to match target lighting. 
We evaluated three models: IC-Light~\cite{zhang2025scaling}, Self-OSR~\cite{yu2020self} and ColorTransfer~\cite{lee2020deep}.
As shown in left of Fig.~\ref{fig:relight}, IC-Light struggled to relight complex outdoor scenes while Self-OSR and ColorTransfer showed only limited performance.
These results indicate that current image-based relighting methods generalize poorly to outdoor urban scenes. Thus, \codename offers promising potential to support future work in image-based relighting for outdoor environments.

\end{document}